\documentclass[lettersize,journal]{IEEEtran}

\usepackage[cmex10]{amsmath}
\usepackage{amssymb}
\usepackage{amsfonts}
\usepackage{array}
\usepackage{textcomp}
\usepackage{stfloats}
\usepackage{url}
\usepackage{verbatim}
\usepackage{cite}
\usepackage{xcolor}
\usepackage{threeparttable}

\usepackage{epsfig}

\usepackage{booktabs}
\usepackage{multirow}
\usepackage{makecell}
\usepackage{stfloats}
\usepackage{float}

\usepackage{bm}
\usepackage{multirow}
\usepackage[Symbol]{upgreek}
\usepackage{marvosym}
\usepackage{amsfonts}

\usepackage{booktabs} 
\usepackage{graphicx} 
\usepackage{amssymb}  
\usepackage{cite}

\usepackage{diagbox}

\usepackage{graphicx}
\usepackage{array}

\usepackage[caption=false,font=footnotesize,labelfont=sf,textfont=sf]{subfig}

\usepackage{stfloats}

\usepackage{hyperref}
\usepackage{multirow}
\usepackage{amssymb}
\usepackage{bm}
\usepackage{mathrsfs}
\usepackage{epstopdf}
\usepackage[dvipsnames]{xcolor}
\definecolor{high}{HTML}{26408B}
\definecolor{mid}{HTML}{81B1D5}
\usepackage{algpseudocode}

\usepackage{amsmath}

\usepackage{multirow}
\usepackage{graphicx}
\usepackage{booktabs}
\usepackage{amssymb}

\begin{document}
\title{Keep It CALM: Toward Calibration-Free Kilometer-Level SLAM with Visual Geometry Foundation Models via an Assistant Eye}
\author{Tianjun~Zhang$^{1}$, Fengyi~Zhang$^{2}$, 	    ~Tianchen~Deng$^{1}$,~\IEEEmembership{Graduate Student Member,~IEEE,}
	~Lin~Zhang$^{3}$,~\IEEEmembership{Senior Member,~IEEE,}
	~Hesheng~Wang$^{1*}$,~\IEEEmembership{Senior Member,~IEEE}
	\IEEEcompsocitemizethanks{
		\IEEEcompsocthanksitem This work was supported in part by the Natural Science Foundation of China under Grant 62225309, Grant U24A20278, Grant 62361166632, Grant U21A20480, and Grant 62503315, the China Postdoctoral Science Foundation under Grant 2025M771737 (Corresponding author: Hesheng Wang).
		\IEEEcompsocthanksitem $^{1}$Tianjun Zhang, Tianchen Deng and Hesheng Wang are with the School of Automation and Intelligent Sensing, and Shanghai Key Laboratory of Navigation and Location Based Services, Shanghai Jiao Tong University, Shanghai 200240, China (email: \{z619850002, dengtianchen, wanghesheng\}@sjtu.edu.cn).
		\IEEEcompsocthanksitem $^{2}$Fengyi Zhang is with the School of Electrical Engineering and Computer Science, The University of Queensland, Brisbane 4067, Australia (email: fengyi.zhang@uq.edu.au). 
		\IEEEcompsocthanksitem $^{3}$Lin Zhang is with the School of Computer Science and Technology, Tongji University, Shanghai 201804, China (email: cslinzhang@tongji.edu.cn).
}}


\markboth{IEEE Transactions on Pattern Analysis and Machine Intelligence}%
{How to Use the IEEEtran \LaTeX \ Templates}

\maketitle


\begin{abstract}
	Visual Geometry Foundation Models (VGFMs) demonstrate remarkable zero-shot capabilities in local reconstruction. However, deploying them for kilometer-level Simultaneous Localization and Mapping (SLAM) remains challenging. In such scenarios, current approaches mainly rely on linear transforms (e.g., Sim3 and SL4) for sub-map alignment, while we argue that a single linear transform is fundamentally insufficient to model the complex, non-linear geometric distortions inherent in VGFM outputs. Forcing such rigid alignment leads to the rapid accumulation of uncorrected residuals, eventually resulting in significant trajectory drift and map divergence.
	To address these limitations, we present \textbf{CAL$^\text{2}$M} (\textbf{\underline{C}}alibration-free \textbf{\underline{A}}ssistant-eye based \textbf{\underline{L}}arge-scale \textbf{\underline{L}}ocalization and \textbf{\underline{M}}apping), a plug-and-play framework compatible with arbitrary VGFMs. Distinct from traditional systems, CAL$^\text{2}$M introduces an ``assistant eye'' solely to leverage the prior of constant physical spacing, effectively eliminating scale ambiguity without any temporal or spatial pre-calibration. Furthermore, leveraging the assumption of accurate feature matching, we propose an epipolar-guided intrinsic and pose correction model. Supported by an online intrinsic search module, it can effectively rectify rotation and translation errors caused by inaccurate intrinsics through fundamental matrix decomposition. Finally, to ensure accurate mapping, we introduce a globally consistent mapping strategy based on anchor propagation. By constructing and fusing anchors across the trajectory, we establish a direct local-to-global mapping relationship. This enables the application of non-linear transformations to elastically align sub-maps, effectively eliminating geometric misalignments and ensuring a globally consistent reconstruction. 
	The source code of CAL$^\text{2}$M will be publicly available at \url{https://github.com/IRMVLab/CALM}.
\end{abstract}


\section{Introduction}
\IEEEPARstart{S}{imultaneous} Localization and Mapping (SLAM) and 3D reconstruction constitute the fundamental pillars of autonomous systems \cite{TPAMI25_UAD, TPAMI25_RoboBEV}, enabling robots and augmented reality devices to perceive and navigate in complex environments. Traditional visual SLAM pipelines, ranging from feature-based methods \cite{ORB-SLAM, ORB-SLAM2, ORB-SLAM3} to direct ones \cite{SVO, LSD-SLAM, DSO}, have achieved varying degrees of maturity. However, their reliance on accurate sensor calibration remains a significant bottleneck. 
In practical deployment, pre-calibration is often tedious and prone to mechanical degradation over time. Moreover, with the rapid evolution of general intelligence, the demand for robustness and flexibility \cite{Autonomous, TPAMI25_MBASLAM, TPAMI25_LN3Diff} often outweighs the pursuit of the absolute localization accuracy in many emerging autonomous applications. Consequently, the development of calibration-free and highly generalizable SLAM systems has become a critical pursuit in the computer vision community.

Recently, the field has witnessed a paradigm shift with the emergence of Visual Geometry Foundation Models (VGFMs). Pioneering works such as DUSt3R \cite{DUSt3R} and MASt3R \cite{Mast3R} have demonstrated that 3D reconstruction can be effectively modeled as a dense, pixel-wise regression task, bypassing the rigid geometric constraints of classical Structure-from-Motion (SfM). Building upon this success, subsequent models like VGGT \cite{VGGT} and Pi3 \cite{Pi3} have further pushed the boundaries to sequence-level reconstruction. By leveraging massive datasets, these VGFMs demonstrate exceptional zero-shot generalization capabilities, enabling robust feature matching and 3D structure recovery from uncalibrated image sequences in challenging scenarios where traditional methods often fail.

Despite these transformative advancements, a significant gap remains between local reconstruction and long-trajectory mapping. Specifically, while existing VGFMs perform admirably within short temporal windows (typically dozens of frames), applying their capability to kilometer-level and long-term trajectories proves non-trivial. The primary obstacle on such applications lies in computational constraints and the models' inherent design, which prioritizes short-term coherence over long-sequence processing. To circumvent these resource bottlenecks, recent approaches, such as VGGT-Long \cite{VGGT-Long} and VGGT-SLAM \cite{VGGT-SLAM}, have adopted a sub-map based strategy, which involves partitioning a long trajectory into shorter segments and stitching them together using linear geometric transformations, such as Sim3 or SL4 transformations.

\begin{figure*}[t]
	\centering
	\includegraphics[width=\linewidth]{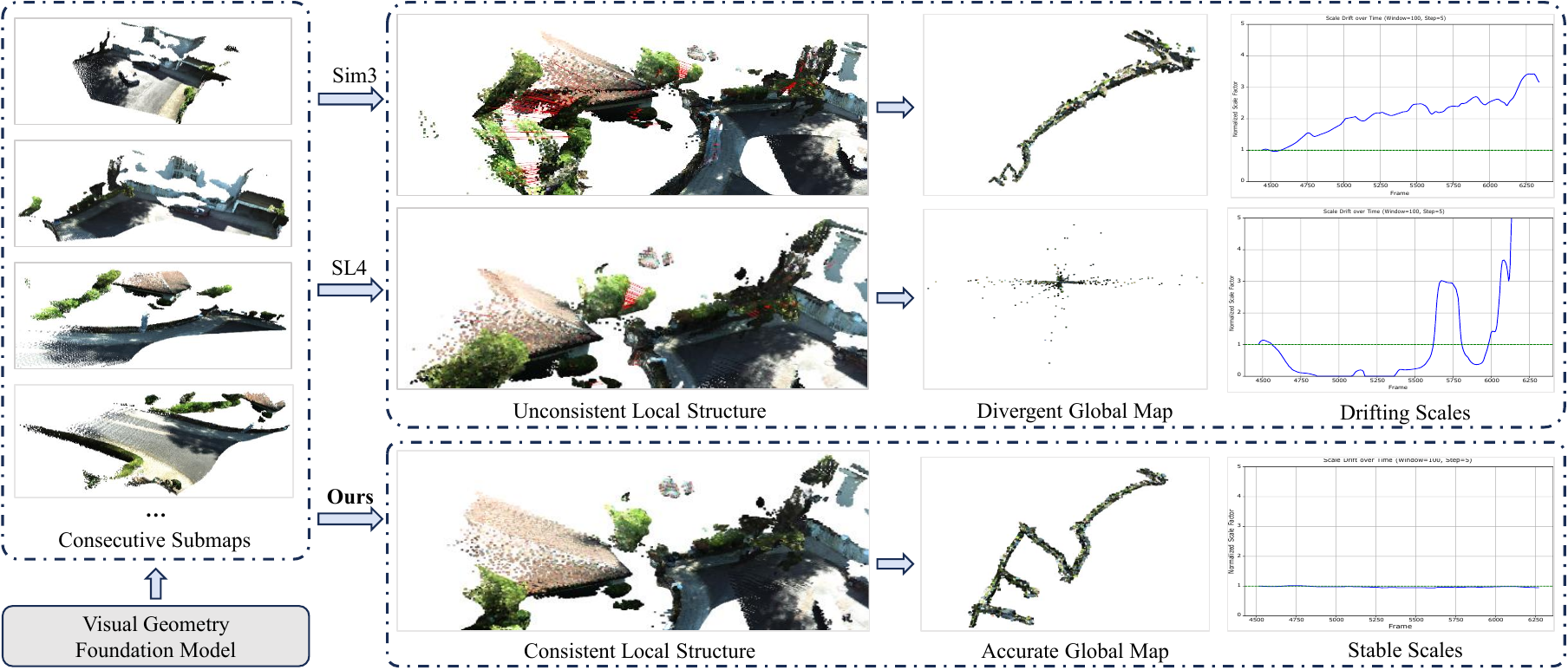}
	\caption{\textbf{Comparison of different VGFM-based incremental localization and mapping methods on kilometer-scale trajectories (without loop closure detection)}. While VGFMs generate locally coherent sub-maps, previous approaches relying on linear alignment (Sim3 or SL4) fail to model the non-linear error distribution, resulting in significant geometric misalignment, global map divergence, and scale drift. In contrast, our CAL$^\text{2}$M pipeline can effectively eliminate these inaccuracies and achieve stable scale recovery, consistent local alignment, and accurate global mapping.}
	\label{fig:motivation}
	\vspace{-2mm}
\end{figure*}

However, we argue that relying solely on linear alignment strategies is insufficient for addressing the complex error distribution inherent in VGFMs. It is important to note that while VGFMs excel at establishing accurate feature correspondences, the dense geometric structure derived from them often suffers from significant inaccuracies. Through detailed analysis, we observe that these geometric inaccuracies are multifaceted and coupled, preventing them from being perfectly rectified by linear transformations. Specifically, these inaccuracies arise from three distinct sources:
\begin{enumerate}
	\item{\textbf{Inherent Scale Ambiguity.} Restricted by the projective nature of the camera observation model, the absolute metric scale is theoretically unobservable from monocular inputs. Consequently, even with semantic priors learned from massive datasets, VGFMs usually cannot recover the accurate metric scale without external geometric constraints, resulting in significant scale fluctuations across different sub-maps.}
	\item{\textbf{Intrinsic and Pose Uncertainty.} VGFMs often struggle with accurate intrinsic estimation due to the affine ambiguity of the camera. For instance, in degenerate motions like straight-line driving, distinguishing between a wide road captured with a large Field of View (FoV) and a narrow road with a small FoV is mathematically ill-posed. The intrinsic inaccuracies will subsequently propagate into pose estimation, causing pose errors in both rotation and translation.}
	\item{\textbf{Non-Linear Structural Distortion.} Finally, based on the accurate feature correspondences while inaccurate intrinsics and poses, the estimated depth maps and 3D point clouds frequently exhibit non-linear warping. When existing methods force-fit these non-linearly distorted sub-maps using linear Sim3 or SL4 transforms, the geometric errors usually are not fully corrected but propagated along with the trajectory as residuals.}
\end{enumerate}
In a sequential alignment framework, where each sub-map is aligned to its predecessor, residuals caused by these inaccuracies can accumulate rapidly, leading to catastrophic long-term trajectory drift and map divergence, as illustrated in Fig.~\ref{fig:motivation}.

To resolve these challenges and bridge the gap between VGFMs and long-trajectory SLAM, we present \textbf{CAL$^\text{2}$M} (\textbf{C}alibration-free \textbf{A}ssistant-eye based \textbf{L}arge-scale \textbf{L}ocalization and \textbf{M}apping). Our proposed framework preserves the flexibility of VGFMs while enforcing global stability. Distinct from traditional synchronized multi-view systems that demand strict hardware synchronization and pre-calibration, CAL$^\text{2}$M introduces an ``assistant eye'' (a secondary camera) solely to leverage the prior of constant physical spacing, which simple yet effective constraint allows us to maintain scale consistency without obviously burdening the system with complex calibration and deployment routines. Our contributions are summarized as follows:

\begin{enumerate}
	\item{\textbf{A VGFM-Driven Calibration-Free SLAM Framework.} We propose a universal, plug-and-play framework capable of integrating with arbitrary VGFMs. By introducing an assistant eye, we utilize the camera spacing prior to fundamentally eliminate scale ambiguity and unify the scale across the trajectory. Our method requires no intrinsic/extrinsic calibration or camera clock synchronization, enabling free and stable kilometer-level localization and mapping.}
	\item{\textbf{Epipolar-Guided Intrinsic and Pose Correction.} We propose a novel strategy to mitigate localization errors caused by the affine ambiguity in the camera observation model. First, we construct an online intrinsic search module to identify accurate camera parameters via confidence scoring on an online-built test bank. Besides, we derive a rigorous pose correction model based on epipolar geometry and fundamental matrix decomposition. Inputing the corrected intrinsics, it can effectively rectify the coupled errors in rotation and translation, ensuring high-accuracy pose estimation.}
	\item{\textbf{Global-Consistent Mapping via Anchor Propagation.} We introduce a globally consistent mapping strategy rooted in the extraction and propagation of geometric anchors. Specifically, we extract reliable geometric anchors within sub-maps and establish a global anchor structure through forward/backward propagation and anchor fusion, thereby constructing a direct local-to-global mapping. Subsequently, we employ non-linear transformations (e.g., Thin Plate Spline) to align sub-maps based on these global anchors, which effectively eliminates geometric inconsistencies between sub-maps and enhances overall mapping accuracy.}
\end{enumerate}

\section{Related Work}
\subsection{Visual 3D Reconstruction}
The recovery of accurate 3D structure from 2D imagery has evolved from rigorous, multi-stage geometric pipelines to end-to-end data-driven neural methods. Structure-from-Motion (SfM) remains the gold standard for sparse reconstruction, with COLMAP \cite{Colmap1, Colmap2} standing as the most prominent framework. By combining robust feature extractors, such as SIFT \cite{SIFT}, with RANSAC-based geometric verification, COLMAP incrementally reconstructs scenes from unordered image collections, refining camera poses and points via global Bundle Adjustment (BA). Once sparse structures are established, multi-view stereo algorithms \cite{SGM, Furukawa2010Stereo} are typically employed to estimate dense depth maps by enforcing photometric consistency across epipolar lines. In recent years, deep learning has significantly advanced this field. Architectures like MVSNet \cite{MVSNet} and its variants \cite{RecurrentMVSNet,  Gu2020Multiview} have replaced hand-crafted matching costs with differentiable 3D cost volumes regularized by 3D CNNs. Such a substitution allows for high-quality depth estimation even in texture-poor regions or non-Lambertian surfaces where traditional methods often fail. Furthermore, efficient variants like PatchmatchNet \cite{PatchMatchNet} have been developed to optimize memory usage using learnable patch-match propagation, enabling high-resolution mapping.

A major paradigm shift occurred with the introduction of NeRF \cite{NeRF}, which proposed representing scenes as continuous neural radiance field. By mapping 5D coordinates (position and viewing direction) to volumetric density and color, NeRF achieved unprecedented photorealistic quality in novel view synthesis. Subsequent improvements rapidly addressed its limitations. For instance, Instant-NGP \cite{InstantNGP} utilized multi-resolution hash encoding to accelerate training orders of magnitude, while Mip-NeRF \cite{MipNeRF} mitigated aliasing artifacts by casting cones rather than rays. PixelNeRF \cite{PixelNeRF}, IBRNet \cite{IBRNet}, and MVSNeRF \cite{MVSNeRF} further extended NeRF \cite{NeRF} by conditioning the radiance fields on input image features, enabling generalization across different scenes. However, a persistent bottleneck for NeRF-based methods is the slow rendering speed caused by the need for dense neural network querying along optical rays to solve the volumetric integral. To alleviate this computational burden, 3D Gaussian Splatting (3DGS) \cite{3DGS} was proposed, replacing implicit neural networks with explicit, anisotropic 3D Gaussian point clouds. By projecting these Gaussians into 2D splats for rasterization, 3DGS achieves real-time rendering speeds. Despite these advances, both NeRF \cite{NeRF} and 3DGS \cite{3DGS} fundamentally rely on the ``per-scene optimization'' paradigm, which implies that they act as ``overfitting'' processes for specific instances rather than general-purpose perception systems. Consequently, the extensive optimization time required for each new scene makes them difficult to adapt for online mapping in long-trajectory scenarios. Moreover, in uncontrolled outdoor environments, these methods tend to geometrically overfit training views, often resulting in structural artifacts or ``floaters'', which also limits their usage in downstream tasks.

\subsection{Visual SLAM}
In contrast to the aforementioned reconstruction methods that often operate in offline batch modes, visual SLAM systems are specifically designed for online, real-time state estimation. Visual SLAM has matured significantly over the past decades, evolving from filter-based to optimization-based backends. Early Monocular SLAM systems, such as MonoSLAM \cite{MonoSLAM}, demonstrated the feasibility of real-time tracking using Extended Kalman Filters. A significant breakthrough came with PTAM \cite{PTAM}, which introduced the parallel tracking and mapping architecture, a design pattern separating real-time pose estimation from expensive map refinement. Such a paradigm has been adopted by most modern systems, such as ORB-SLAM \cite{ORB-SLAM} and SVO \cite{SVO}. While feature-based methods focused on minimizing reprojection error of keypoints, direct methods like LSD-SLAM \cite{LSD-SLAM} and DSO \cite{DSO, LDSO} operated directly on pixel intensities to generate semi-dense maps, exploiting photometric consistency to track pixels even in low-texture areas. Nevertheless, a fundamental limitation of all monocular methods is the inherent scale ambiguity. To resolve this, stereo SLAM systems were developed to recover metric scale via a fixed baseline. S-PTAM \cite{SPTAM} adapted the parallel architecture for stereo setups, while ORB-SLAM2 \cite{ORB-SLAM2} and ORB-SLAM3 \cite{ORB-SLAM3} provided stereo modes that effectively prevent scale drift by triangulating points from calibrated stereo pairs. However, stereo systems depend heavily on rigorous pre-calibration and rigid mechanical structures, even little mechanical deformations or thermal expansion can invalidate the calibration, leading to significant depth inaccuracies \cite{StereoLSD}.

In recent years, Learning-based SLAM has emerged as a robust alternative to traditional geometry-based pipelines. DROID-SLAM \cite{DROID-SLAM} revolutionized the field by formulating SLAM as a dense bundle adjustment problem within a deep learning framework. 
To improve computational efficiency, DPVO \cite{DPVO, DPV-SLAM} and TartanVO \cite{TartanVO} utilized sparse patch-based tracking, treating patches as trackable entities. Parallel to these explicit methods, research into neural implicit SLAM has also flourished. Early works like iMAP \cite{IMAP} and NICE-SLAM \cite{NICE-SLAM} demonstrated the potential of optimizing implicit neural fields online. Loopy-SLAM \cite{LoopySLAM} extended these implicit representations by incorporating robust loop closure detection and global optimization to correct accumulated drift over longer sequences. 
Additionally, MNE-SLAM \cite{deng2025mne} further expanded the scope of neural SLAM by enabling collaborative mapping in multi-agent mobile robot systems. 
In parallel, the success of 3DGS has inspired 3DGS-based SLAM systems like SplaTAM \cite{SplaTAM}, GS-SLAM \cite{GS-SLAM}, and MonoGS \cite{MonoGS}, which leverage Gaussians for high-fidelity dense mapping. 
While NeRF-based methods and 3DGS-based ones can achieve high rendering quality, compared to feed-forward models, these optimization-based methods are significantly slower in mapping updates due to the need for continuous gradient backpropagation. 


\subsection{Visual Geometry Foundation Models}
The emergence of VGFMs represents a transformative shift toward calibration-free, generalizable 3D perception. DUSt3R \cite{DUSt3R} pioneered this direction by formulating unconstrained stereo reconstruction as a direct pointmap regression task. Instead of explicit triangulation, it predicts dense 3D coordinates in the camera frame, implicitly solving for intrinsics and extrinsics. MASt3R \cite{Mast3R} enhanced such capability by grounding image matching in 3D space, employing a coarse-to-fine matching strategy that significantly improves robustness under extreme viewpoint changes. To address the efficiency bottlenecks in these models, Fast3R \cite{Fast3R} optimized the architecture to accelerate inference speeds for real-time applications. Building upon these pairwise foundations, models like MUSt3R \cite{MUSt3R} and Spann3R \cite{Span3R} explored multi-view consistency and sparse-view generalization, aggregating information from multi-angle to improve reconstruction coherence. Subsequently, to capture temporal consistency in continuous video streams, VGGT \cite{VGGT} introduced Transformer-based architectures, treating video frames as tokens in a temporal context window. 
As a pioneering work in the landscape of visual geometry foundation models, VGGT \cite{VGGT} has inspired a wide array of subsequent variants that build upon its architecture, extending its capabilities to other visual tasks \cite{deng2025best3dscenerepresentation, deng2025relocvggtvisualrelocalizationgeometry, unipr-3d}. 
Similarly, Pi3 \cite{Pi3} integrated probabilistic priors to model the generation uncertainty, and MapAnything \cite{MapAnything} proposed a universal feed-forward metric reconstruction framework to handle diverse real-world inputs.

Despite their impressive zero-shot capabilities in short sequences, applying VGFMs to kilometer-level long trajectories remains an open challenge. Initial efforts, such as SLAM3R \cite{SLAM3R} and MASt3R-SLAM \cite{Mast3rSLAM}, integrate DUSt3R \cite{DUSt3R} and MASt3R \cite{Mast3R} into SLAM frameworks, while CUT3R \cite{CUT3R} employs a divide-and-conquer strategy to scale up reconstruction. However, constrained by the performance limitations of their backbone models, they are highly susceptible to tracking loss and trajectory divergence, particularly in complex outdoor environments. To address the stability issues, VGGT-Long \cite{VGGT-Long} and VGGT-SLAM \cite{VGGT-SLAM} leverage the more powerful model, VGGT \cite{VGGT}. Since VGGT \cite{VGGT} supports sequence-level input, these methods adopt a sub-map based strategy, partitioning long trajectories into chunks and stitching the resulting sub-maps using Sim3 or SL4 transformations. Nevertheless, these methods largely rely on the assumption that sub-maps can be aligned via linear transformations, which overlooks the non-linear distortions and coupled intrinsic-pose errors inherent in VGFMs. 
Consequently, these errors usually propagate rapidly, leading to severe localization drift and mapping divergence.

\section{Preliminaries and Problem Formulation}
Visual Geometry Foundation Models (VGFMs) have evolved from pair-wise pointmap regressors to powerful sequence-level estimators capable of processing an arbitrary set of uncalibrated images. Modern architectures, such as VGGT \cite{VGGT}, Pi3 \cite{Pi3}, and MapAnything \cite{MapAnything}, leverage Transformer-based backbones to model global geometric consistency across a context window.
Formally, let $\Phi$ denote a pre-trained VGFM. Given an input of $N$ images $\mathcal{I} = \{\mathbf{I}_1, \mathbf{I}_2, \dots, \mathbf{I}_N\}$, where $\mathbf{I}_i \in \mathbb{R}^{H \times W \times 3}$, the model directly infers the set of intrinsics, camera poses, depth maps, and confidence maps for the entire sequence,
\begin{equation}
	(\mathcal{K}, \mathcal{T}, \mathcal{D}, \mathcal{C}) = \Phi(\mathcal{I}).
\end{equation}
Here, $\mathcal{K} = \{\mathbf{K}_1, \dots, \mathbf{K}_N\}$ denotes the intrinsic matrices, which may vary per frame even for a single camera to account for affine ambiguities. $\mathcal{T} = \{\mathbf{T}_1, \dots, \mathbf{T}_N\}$ represents the estimated camera poses in $\text{SE}(3)$. $\mathcal{D} = \{\mathbf{D}_1, \dots, \mathbf{D}_N\}$ constitutes the dense depth maps, and $\mathcal{C} = \{\mathbf{C}_1, \dots, \mathbf{C}_N\}$ contains the confidence maps quantifying the pixel-wise uncertainty of the geometric estimation.
Unlike traditional SfM relying on iterative BA, these models output geometry in a single forward inference, producing locally consistent reconstruction results.

Based on VGFMs, we aim to address the problem of kilometer-level SLAM using an uncalibrated dual-camera setup. Let the video streams from the primary and assistant cameras be denoted as $\mathcal{V}_{pri} = \{\mathbf{I}_{1}^{p}, \mathbf{I}_{2}^{p}, \dots\}$ and $\mathcal{V}_{ast} = \{\mathbf{I}_{1}^{a}, \mathbf{I}_{2}^{a}, \dots\}$. Our CAL$^\text{2}$M operates under following settings:
\begin{enumerate}
	\item \textbf{Calibration-Free Setup:} The ground truth intrinsics $\mathbf{K}_{gt}$ and the extrinsic transformation $\mathbf{T}_{ext}$ are unknown. The relative rotation is unconstrained, provided that the primary and assistant cameras maintain a sufficient overlapping FoV to enable inter-view feature matching.
	\item \textbf{Constant Spacing Assumption:} The specific value of the physical baseline distance $d$ is not strictly required to be known (if absolute scale is not required). However, the algorithm assumes the translation norm $\|\mathbf{t}_{pri \to ast}\|$ remains invariant throughout the operation, serving as a consistent reference for scale unification.
	\item \textbf{Loose Synchronization:} We assume the streams share a common timeline where images in two streams can be associated via timestamps, without requiring strict hardware trigger synchronization. It's worth noting that, for the sake of clarity, in most of the subsequent algorithm explanations, we first explain the algorithms based on a synchronized clock setting, and then, where necessary, explain how to extend them to the asynchronous setting.
\end{enumerate}
The objective of CAL$^\text{2}$M is to estimate a globally consistent, drift-free trajectory $\mathcal{T}_{global}$ and map $\mathcal{M}_{global}$ by effectively chaining the local predictions of $\Phi$.

\begin{figure*}[t]
	\centering
	\includegraphics[width=\linewidth]{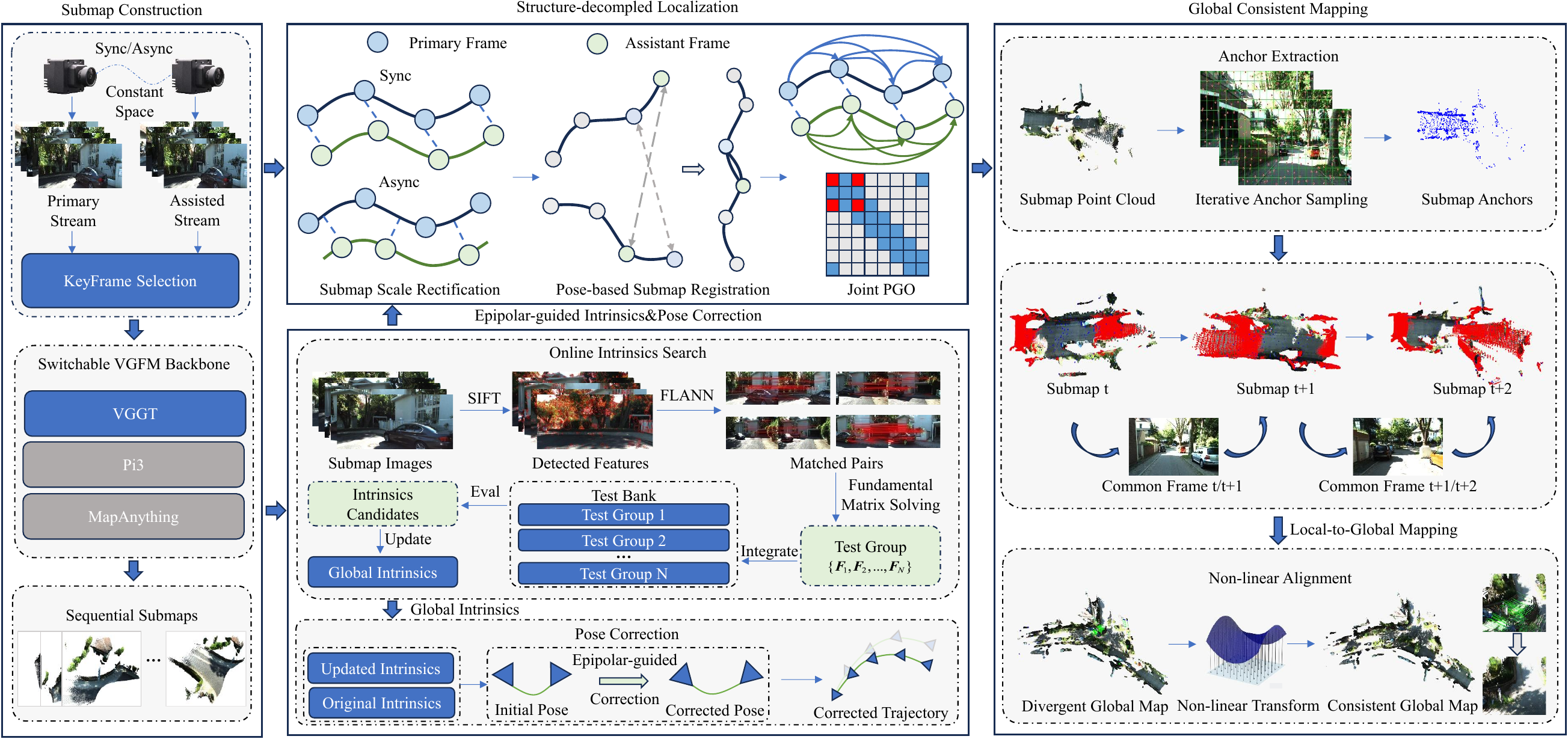}
	\caption{\textbf{System architecture of CAL$^\text{2}$M.} The pipeline begins by constructing sub-maps from the primary and assistant image streams using a plug-and-play VGFM backbone. The raw estimates undergo epipolar-guided intrinsic search and pose correction. Then, the scale of different submaps are unified via the constant spacing prior, and the global trajectory is optimized by our Joint Primary-Assistant PGO. Finally, the optimized trajectory and raw point clouds are processed by our globally consistent mapping module, which utilizes anchor propagation and non-linear transform to achieve seamless reconstruction.}
	\label{fig:pipeline}
	\vspace{-2mm}
\end{figure*}

\section{The CAL$^\text{2}$M Pipeline}
The overall pipeline of CAL$^\text{2}$M is illustrated in Fig.~\ref{fig:pipeline}. Upon receiving the image streams $\mathcal{V}_{pri}$ and $\mathcal{V}_{ast}$, the system first executes a keyframe selection strategy. The selected keyframes are then grouped into temporal batches and fed into the VGFM $\Phi$ to estimate local sub-map information.  Subsequently, sub-maps are then scale-aligned via the constant spacing prior and refined by our Primary-Assistant Joint Pose Graph Optimization (PGO). Distinctively, our backend relies exclusively on pose and loop closure constraints, effectively decoupling trajectory estimation from dense structural noise. 

Furthermore, to address the specific limitations of VGFMs, the pipeline integrates two novel refinement modules: an epipolar-guided intrinsic and pose correction module designed to mitigate localization errors stemming from affine ambiguity of the observation model, and a global-consistent mapping module based on anchor propagation to enhance the accuracy and global consistency of the final dense mapping.
These two modules are introduced in detail in Sec. \ref{sec:correction} and \ref{sec:mapping}, respectively.

%

\subsection{Sub-map Construction}
\label{subsec:submap_construction}

To efficiently process long video streams, we partition the trajectory using a sliding-window strategy. Motivated by \cite{VGGT-SLAM}, primary keyframes are selected based on optical flow \cite{ShiTomasi, LK} to ensure sufficient parallax: a new frame is triggered only when the average disparity exceeds $\tau_{flow}$, continuing until the batch size reaches $N_{max}$. Corresponding assistant views are then associated. For synchronized streams, we retrieve timestamp-matched frames, while for asynchronous scenarios, we identify the temporally closest frame and apply de-duplication. The union of these frames is fed into VGFM $\Phi$ to infer local geometry ($\mathcal{K}_{sub}, \mathcal{T}_{sub}, \mathcal{D}_{sub}, \mathcal{C}_{sub}$). To maintain geometric continuity, we retain the last $N_{overlap}$ frames as a connecting bridge for the initialization of the next sub-map.

\subsection{Scale Rectification and Sub-map Alignment}
\label{subsec:scale_alignment}

Although VGFMs infer locally consistent structures, the scale of each sub-map varies due to the monocular scale ambiguity. We leverage the constant physical spacing between the primary and assistant cameras as a geometric invariant to unify the scales across the entire trajectory.

Let $\mathcal{S}_k$ denote the $k$-th sub-map. For each pair of primary and assistant views $(i, j)$ within $\mathcal{S}_k$, the VGFM estimates their poses as $\mathbf{T}_{i, k}^p, \mathbf{T}_{j, k}^a$. The instantaneous camera spacing estimated by the model is the Euclidean distance between their optical centers,
\begin{equation}
	d_{i, k} = \| \mathbf{p}(\mathbf{T}_{i, k}^p) - \mathbf{p}(\mathbf{T}_{j, k}^a) \|_2 ,
\end{equation}
where $\mathbf{p}(\cdot)$ extracts the translation component from the pose matrix. In the asynchronous mode, the assistant frame at timestamp $t_{ast}$ may not have an exactly matching primary frame. We address this by interpolating the primary camera pose. 
The interpolated reference pose is computed via $\text{SE}(3)$ interpolation, and the spacing $d_{i, k}$ is then computed between the interpolated $\mathbf{T}_{ref}^p$ and the assistant pose $\mathbf{T}_{j, k}^a$.

We compute the representative spacing $\bar{d}_k$ for sub-map $\mathcal{S}_k$ by averaging $d_{i, k}$ over all valid pairs. Using the average spacing of the first sub-map $\bar{d}_{1}$ as the global reference, the scale rectification factor $s_k$ is derived as $s_k = \bar{d}_{1} / \bar{d}_k$. We apply this factor to rectify the translation vectors and depth maps within the sub-map $\mathcal{S}_k$. Crucially, since $s_k$ is derived solely from the current observations and a fixed global reference, this process effectively decouples the scale estimation from previous sub-maps, thereby thoroughly eliminating scale drift.

Following scale rectification, we initialize the pose $ \mathbf{T}_{k \to w}$ of the current sub-map in the global world coordinate system. We utilize the first common frame $f$ to calculate $\mathbf{T}_{k \to w}$ as, 
\begin{equation}
	\mathbf{T}_{k \to w} = \mathbf{T}_{w, f} \cdot (\mathbf{T}_{k, f})^{-1} .
\end{equation}
All poses in $\mathcal{S}_k$ are then transformed to the world frame via $\mathbf{T}_{world} = \mathbf{T}_{k \to w} \cdot \mathbf{T}_{local}$. It is worth noting that our alignment relies exclusively on camera poses rather than dense point clouds. This design choice is motivated by the observation that VGFMs generally perform more robust in pose estimation compared to dense structure recovery, particularly in large-scale open outdoor environments. Even when estimated poses are relatively accurate, the dense point clouds often exhibit obvious non-linear distortions. Consequently, a pose-based alignment strategy provides a more stable and robust sub-map alignment compared to structure-based registration.

\subsection{Primary-Assistant Joint PGO}
\label{subsec:joint_pgo}

To ensure accurate localization, upon specific events (successful loop closure detection and global-level updates from the pose correction module), CAL$^\text{2}$M triggers a global backend optimization to refine the state vector $\boldsymbol{\Theta} = \{ \mathbf{T}_{w, 1}^p, \dots, \mathbf{T}_{w, N}^p, \mathbf{T}_{ext} \}$, comprising the global primary poses and the static extrinsic transformation. A unique feature of our system is the implicit modeling of the assistant trajectory: rather than optimizing assistant poses independently, we parameterize them as $\mathbf{T}_{w, i}^a = \mathbf{T}_{w, i}^p \cdot \mathbf{T}_{ext}$, thereby better utilizing the prior of the fixed extrinsics.

The optimization minimizes the Mahalanobis distances of primary, assistant, prior, and loop closure constraints ($E_{pri}$, $E_{ast}$, $E_{prior}$, and $E_{loop}$). For the primary trajectory, we construct dense odometry constraints between frame $i$ and its preceding $K$ frames using scale-rectified VGFM measurements. For the assistant trajectory, we employ a universal coupled constraint formulation. The relative measurement $\Delta \mathbf{T}_{j, i}^{ast}$ can be directly derived from VGFM outputs in synchronous mode. In asynchronous scenarios, it is computed via SE(3) interpolation of the assistant trajectory at primary timestamps $t_i$ and $t_j$. Finally, the measurement $\mathbf{T}_{j, i}^{ast}$ forms the residual,
\begin{equation}
	\begin{split}
		E_{ast}(i, j) & = \big\| \log \big( (\Delta \mathbf{T}_{j, i}^{ast})^{-1}  \cdot (\mathbf{T}_{w, j}^a)^{-1} \mathbf{T}_{w, i}^a \big) \big\|_{\boldsymbol{\Sigma}_{ast}}^2  \\
		& \text{where } \mathbf{T}_{w, k}^a = \mathbf{T}_{w, k}^p \mathbf{T}_{ext} .
	\end{split}
\end{equation}

To prevent parameter divergence under degenerate motions (e.g., constant velocity), a soft prior constraint $E_{prior}$ is further applied to $\mathbf{T}_{ext}$, anchored to the average relative pose estimated from the first sub-map. Additionally, loop closure constraints $E_{loop}$ are integrated upon detection. This non-linear optimization problem is solved using the Levenberg-Marquardt algorithm within the GTSAM framework \cite{GTSAM}.


\subsection{Loop Closure Detection}
\label{subsec:loop_closure}

To eliminate long-term drift and ensure global consistency, we incorporate the loop closure detection mechanism. Motivated by \cite{VGGT-Long, VGGT-SLAM}, before processing a new sub-map with the VGFM, we extract descriptors for the incoming keyframes using the SALAD model \cite{SALAD}. These descriptors are compared against historical keyframes using Euclidean distance to identify the most similar candidate pairs between the current view and visited locations.
Once a potential loop closure is detected, we leverage the zero-shot generalization capability of the VGFM for constraint generation. Instead of relying on traditional feature matching or PnP solvers, we append the retrieved historical image to the input batch of the current sub-map. By feeding this augmented batch into the VGFM, the model naturally infers the geometric relationship between the current frame and the historical one within a single forward pass. This yields a long-term relative pose measurement $\mathbf{T}_{loop}$, which is subsequently integrated into the backend PGO as the loop constraint $E_{loop}$.

\section{Epipolar-Guided Intrinsic and Pose Correction}
\label{sec:correction}

As aforementioned, the geometric reliability of VGFMs is often compromised by affine ambiguity, where inaccurate intrinsic estimation inevitably induces systematic pose deviations. To address this, leveraging the high accuracy of the model's feature correspondences, we present a unified rectification framework based on epipolar geometry. Specifically, we first introduce an online intrinsic search method (Sec.~\ref{subsec:intrinsic_search}) to recover optimal intrinsics by minimizing epipolar errors over an online-constructed test bank. After that, via the fundamental matrix decomposition, we derive a pose correction model (Sec.~\ref{subsec:pose_model}) that can effective correct the local poses in each sub-map estimation using the global intrinsics. Practical integration details are provided in Sec.~\ref{subsec:correction_implementation}.

\subsection{Online Intrinsics Search}
\label{subsec:intrinsic_search}

The primary objective of this module is to identify optimal intrinsics by filtering the fluctuating VGFM estimations. To achieve this, we online-construct a test bank rooted in epipolar geometry. Specifically, for every five processed sub-maps, we initiate the construction of a new test group by extracting SIFT features \cite{SIFT} and then match features among images in a pairwise manner to establish multiple image pairs. Based on these feature matches, we compute the fundamental matrix $\mathbf{F}$ for each image pair. A test group is successfully instantiated only if it contains at least $N_{pair}$ qualified image pairs, with each pair sharing more than $N_{feature}$ reliable feature matches. Furthermore, to balance computational load, the total capacity of the test bank is limited to $N_{group}$ groups.

With the test bank established, we quantify the confidence of an intrinsic candidate $\mathbf{K}$ based on the spectral properties of the essential matrix $\mathbf{E} = \mathbf{K}^T \mathbf{F} \mathbf{K}$. Ideally, $\mathbf{E}$ should have singular values satisfying $\sigma_1 = \sigma_2$ and $\sigma_3 = 0$. Standing on this point, we design a scoring function as,
\begin{equation}
	Score(\mathbf{K}, \mathbf{F}) = \frac{|\sigma_1 - \sigma_2|}{\sigma_1} + \frac{|\sigma_3|}{\sigma_1} .
\end{equation}
The global confidence of $\mathbf{K}$ is then calculated as the average score across all matrices currently stored in the bank. Lower score represents better intrinsics.

In the operation phase, we maintain a global intrinsic matrix $\mathbf{K}_{global}$, initialized from the first sub-map. The update strategy proceeds in two stages: (1) whenever a new sub-map is generated, we update $\mathbf{K}_{global}$ if the new estimate yields a lower score; (2) upon adding a new test group, we re-evaluate all historical estimates against the updated bank to select the global minimum. Our strategy can effectively fuse VGFM priors with sparse geometric verification, filtering out erroneous intrinsic estimates caused by affine ambiguity.

\subsection{Pose Correction Model}
\label{subsec:pose_model}

With the global intrinsics $\mathbf{K}$ via the online search, our goal is to rectify the poses derived from the erroneous intrinsics $\mathbf{K}_{est}$ output by the VGFM. We propose a correction model rooted in the invariance of the fundamental matrix.
Let the true relative rotation and translation be $\mathbf{R}$ and $\mathbf{t}$. The VGFM, utilizing $\mathbf{K}_{est}$, produces the biased estimates $\mathbf{R}_{est}$ and $\mathbf{t}_{est}$. We define the scaling matrix $\mathbf{S}$ and the deviation matrix $\boldsymbol{\Delta}$,
\begin{equation}
	\mathbf{S} = \mathbf{K}_{est} \mathbf{K}^{-1} = \text{diag}(s_x, s_y, 1), \quad \boldsymbol{\Delta} = \mathbf{S} - \mathbf{I} .
\end{equation}
The relationship between the true essential matrix $\mathbf{E}$ and the estimated one $\mathbf{E}_{est}$ is given by,
\begin{equation}
	\mathbf{E} = \mathbf{K}^T \mathbf{F} \mathbf{K} = \mathbf{S}^{-1} \mathbf{E}_{est} \mathbf{S}^{-1} .
	\label{equ:essential_decomp}
\end{equation}

\subsubsection{Translation Correction}
We first address the translation component. Substituting the essential matrix definition $\mathbf{E} = [\mathbf{t}]_{\times} \mathbf{R}$ into Eq. \ref{equ:essential_decomp} yields $[\mathbf{t}]_{\times} \mathbf{R} = \mathbf{S}^{-1} ([\mathbf{t}_{est}]_{\times} \mathbf{R}_{est}) \mathbf{S}^{-1}$. To isolate the translation, we theoretically assume the rotation estimate is consistent ($\mathbf{R} \approx \mathbf{R}_{est}$). This leads to the general matrix form of the translation distortion,
\begin{equation}
	[\mathbf{t}]_{\times} \approx \mathbf{M} = \mathbf{S}^{-1} [\mathbf{t}_{est}]_{\times} \mathbf{R}_{est} \mathbf{S}^{-1} \mathbf{R}_{est}^{-1}.
\end{equation}
To recover the translation vector $\mathbf{t}$, we can extract the relevant skew-symmetric elements from $\mathbf{M}$.
Furthermore, in practical video processing where inter-frame rotation is usually small ($\mathbf{R}_{est} \approx \mathbf{I}$), and such a model can be explicitly simplified as,
\begin{equation}
	\mathbf{t}_{raw} \approx \left[ \frac{1}{s_y} t_{est}^x, \quad \frac{1}{s_x} t_{est}^y, \quad \frac{1}{s_x s_y} t_{est}^z \right]^T .
\end{equation}
However, directly applying this vector would alter the scale of the sub-map by a factor of $1/(s_x s_y)$, leading to inconsistency with the VGFM's output. To preserve scale consistency with the estimated depth, we align the z-axis of the corrected translation by normalizing the vector with a factor of $s_x s_y$,
\begin{equation}
	\mathbf{t} = s_x s_y \cdot \mathbf{t}_{raw}  = \mathbf{S} \cdot \mathbf{t}_{est} .
\label{equ:translation_model_final}
\end{equation}

\subsubsection{Rotation Correction}
Correcting the rotation also involves decoupling the essential matrix. We employ a first-order perturbation analysis. Approximating $\mathbf{S}^{-1} \approx \mathbf{I} - \boldsymbol{\Delta}$, the true essential matrix is approximated as,
\begin{equation}
\begin{aligned}
	\mathbf{E} &\approx \mathbf{E}_{est} - \boldsymbol{\Delta} \mathbf{E}_{est} - \mathbf{E}_{est} \boldsymbol{\Delta}  \\
	&\approx [\mathbf{t}_{est}]_\times \mathbf{R}_{est}  - \boldsymbol{\Delta} [\mathbf{t}_{est}]_\times \mathbf{R}_{est} - [\mathbf{t}_{est}]_\times \mathbf{R}_{est} \boldsymbol{\Delta} .
\label{equ:rotation_decomp1}
\end{aligned}
\end{equation}
Next, we define the corrected rotation $\mathbf{R}$ via a left-multiplication perturbation in the global frame. Let $\boldsymbol{\Theta}$ be the rotation error vector derived from the estimated intrinsics such that the true rotation satisfies $\mathbf{R} \approx (\mathbf{I} - [\boldsymbol{\Theta}]_\times) \mathbf{R}_{est}$. Using the approximated translation $\mathbf{t} \approx (\mathbf{I} + \boldsymbol{\Delta}) \mathbf{t}_{est}$, we derive the perturbation equation for $\mathbf{E} = [\mathbf{t}]_\times \mathbf{R}$,
\begin{equation}
	\mathbf{E} \approx [\mathbf{t}_{est}]_\times \mathbf{R}_{est} + [\boldsymbol{\Delta} \mathbf{t}_{est}]_\times \mathbf{R}_{est} - [\mathbf{t}_{est}]_\times [\boldsymbol{\Theta}]_\times \mathbf{R}_{est} .
	\label{equ:rotation_perturbation}
\end{equation}
Equating Eq. \ref{equ:rotation_perturbation} to Eq. \ref{equ:rotation_decomp1} and simplifying, we have,
\begin{equation}
	\begin{split}
		- [\mathbf{t}_{est}]_\times [\boldsymbol{\Theta}]_\times \mathbf{R}_{est} &+ [\boldsymbol{\Delta} \mathbf{t}_{est}]_\times \mathbf{R}_{est} \\
		\approx - \boldsymbol{\Delta} [\mathbf{t}_{est}]_\times \mathbf{R}_{est} &- [\mathbf{t}_{est}]_\times \mathbf{R}_{est} \boldsymbol{\Delta} .
	\end{split}
\end{equation}
To isolate $\boldsymbol{\Theta}$, we right-multiply by $\mathbf{R}_{est}^T$,
\begin{equation}
	[\mathbf{t}_{est}]_\times [\boldsymbol{\Theta}]_\times = [\boldsymbol{\Delta} \mathbf{t}_{est}]_\times + \boldsymbol{\Delta} [\mathbf{t}_{est}]_\times + [\mathbf{t}_{est}]_\times (\mathbf{R}_{est} \boldsymbol{\Delta} \mathbf{R}_{est}^T) .
	\small
\end{equation}
Applying both sides to the vector $\mathbf{t}_{est}$ and using the vector triple product property, the problem can be converted to a simple linear equation.
Noting that the intrinsic error imposes no constraint on rotation about the translation axis since $\mathbf{t}_{est}$ spans the zero space of the equation, we enforce orthogonality ($\mathbf{t}_{est}^T \boldsymbol{\Theta} \approx 0$) and derive the closed-form solution,
\begin{equation}
	\boldsymbol{\Theta} = \frac{1}{\|\mathbf{t}_{est}\|^2} [\mathbf{t}_{est}]_\times \left( (\mathbf{R}_{est} \boldsymbol{\Delta} \mathbf{R}_{est}^T - \boldsymbol{\Delta}) \mathbf{t}_{est} \right),
\end{equation}
and the corrected rotation matrix can be given as,
\begin{equation}
	\mathbf{R} = \text{Rodrigues}(\boldsymbol{\Theta})^\top \mathbf{R}_{est} .
\label{equ:rotation_model_final}
\end{equation}

\subsection{Adaptive Trajectory Rectification}
\label{subsec:correction_implementation}

The derived correction models are integrated into the CAL$^\text{2}$M pipeline via an incremental strategy. Specifically, for the primary trajectory, we rectify the relative motion between consecutive frames. Let $\mathbf{T}_{i-1, i}^{est}$ be the VGFM-estimated motion. We apply the correction models (Eq. \ref{equ:translation_model_final} and Eq. \ref{equ:rotation_model_final}) to obtain $\mathbf{T}_{i-1, i}^{corr}$ and reconstruct the global trajectory by chaining these poses: $\mathbf{T}_{w, i}^{p} = \mathbf{T}_{w, i-1}^{p} \cdot \mathbf{T}_{i-1, i}^{corr}$. For the assistant trajectory, to enhance stability, we compute the scaling matrix $\mathbf{S}_{joint}$ as the arithmetic mean of the primary and assistant scaling matrices, and then the pose of each frame can be computed by $\mathbf{S}_{joint}$ and its reference frame in the primary trajectory. For async case, its reference frame can be obtained by interpolating the rectified primary trajectory.

To prevent over-correction from inaccurate global intrinsics, we introduce an adaptive damping factor $\lambda \in [0, 1]$ to modulate the error matrix: $\boldsymbol{\Delta}' = \lambda \boldsymbol{\Delta}$. The factor $\lambda$ is determined by the density of the test bank, calculated as $\bar{n} = N_{total} / N_{group}$, where $N_{total}$ is the current number of fundamental matrices and $N_{group}$ is the bank's capacity. We formulate $\lambda$ as,
\begin{equation}
	\lambda = \text{clamp}\left( {\lfloor \bar{n} /({N_{pair} * N_{feature}) \rfloor_{0.1}}}, \quad 0, \quad 1 \right) ,
\end{equation}
where the operator $\lfloor \cdot \rfloor_{0.1}$ denotes rounding down to the nearest multiple of 0.1. Such an adpative correction scales correction strength with geometric confidence. It's worth noticing that, the quantization mechanism is introduced mainly for preventing  frequent global trajectory updates.

\begin{figure}[t]
	\centering
	\subfloat[]{
		\includegraphics[width=0.37\columnwidth]{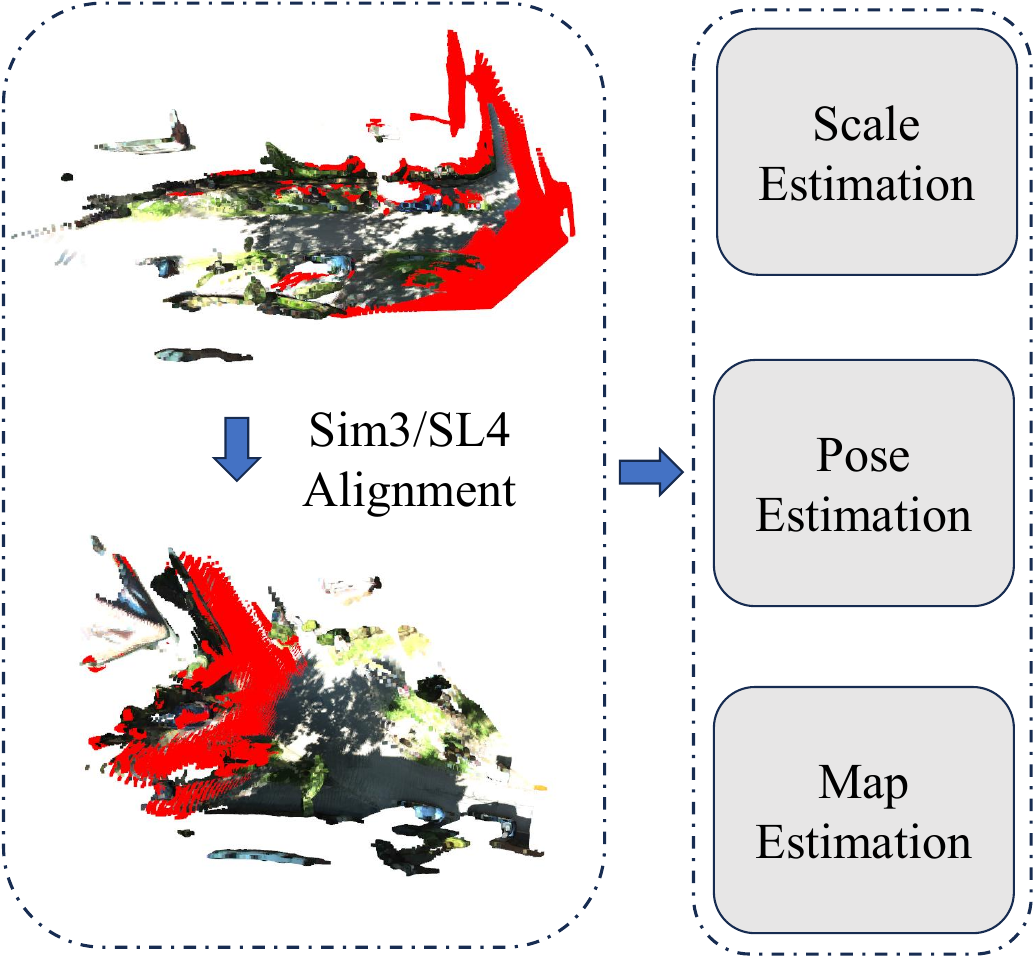}
	}
	\subfloat[]{
		\includegraphics[width=0.58\columnwidth]{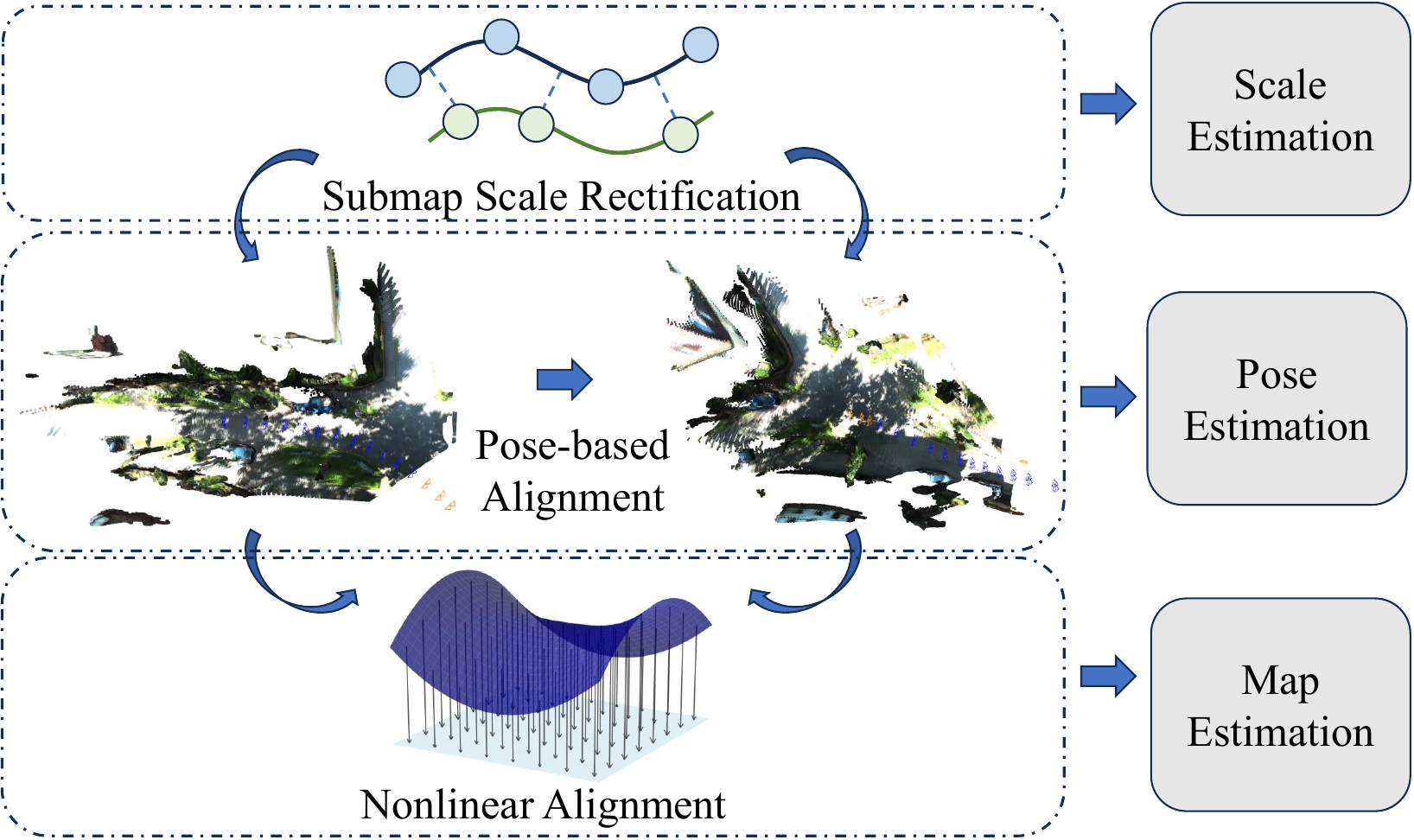}
	}
	\caption{Comparison of different mapping paradigms. (a) Existing methods usually couple the estimation on scale, pose, and structure, causing cumulative non-linear errors. (b) CAL$^\text{2}$M decouples these components, utilizing non-linear anchor alignment to ensure robust global consistency.}
	\label{fig:mapping_concept}
	\vspace{-3mm}
\end{figure}

\section{Anchor-based Global-Consistent Mapping}
\label{sec:mapping}


With the optimized global trajectory, a naive global map can be constructed by rigidly transforming each local sub-map into the world coordinate system. However, this straightforward accumulation typically results in significant structural artifacts, since adjacent sub-maps often exhibit inconsistent 3D structures for the same physical region due to the non-linear depth distortions inherent in VGFMs. 
To address such a challenge, we present a mapping strategy grounded in the high reliability of feature correspondences. Specifically, we extract stable geometric ``anchors'' and propagate them bi-directionally to establish a dense, globally consistent skeletal structure. These anchors can define a direct mapping from the local distorted sub-map to the global reference. Using these anchors as control points, non-linear transforms can be fitted to elastically align each submap. In our implementations, we employ Thin Plate Spline (TPS) \cite{TPS} transform. 
A conceptual comparison highlighting the fundamental difference between our global-consistent mapping strategy and linear-transform based approaches is illustrated in Fig.~\ref{fig:mapping_concept}.

\subsection{Geometric Anchor Extraction}
\label{subsec:anchor_extraction}

Grounded in the premise that VGFMs usually yield accurate feature correspondences, we extract a set of reliable geometric anchors to serve as the control skeleton for the sub-map. These anchors represent physical 3D points that are consistently observed with high confidence across multiple frames.

\subsubsection{Grid-based Initialization and Verification}
Given a sub-map $\mathcal{S}_k$, we initiate the anchor-extraction process from the first frame. To ensure a uniform spatial distribution, we partition the image domain into a fixed grid of size $N_{grid} \times {N}_{grid}$. Within each grid cell, we randomly sample a candidate pixel $\mathbf{u}$ satisfying a high confidence threshold $C(\mathbf{u}) > \tau_{conf}$. Pixel $\mathbf{u}$ is then back-projected into the sub-map's 3D coordinate system using the VGFM-estimated depth $\mathbf{D}(\mathbf{u})$ and the intrinsics $\mathbf{K}$, forming a candidate anchor $\mathbf{P}_{local}$.


Subsequently, we project $\mathbf{P}_{local}$ onto other frames within the sub-map to verify its consistency. A projection is deemed valid only if the absolute difference between the projected depth and the corresponding VGFM-estimated depth lies within a margin proportional to the depth value, scaled by the ratio $\eta_{proj}$. This geometric verification ensures that the anchors are not occlusion boundary points or outliers.

\subsubsection{Iterative Densification}
To maximize scene coverage, we adopt an iterative densification strategy. After processing the anchors from the first frame, we proceed sequentially to the subsequent frames. For the $t$-th frame, we project all previously established anchors onto its image plane. We then identify the grid cells that contain no projected anchors (i.e., regions corresponding to newly observed areas or previously sparse textures). In these empty grids, we sample new candidate anchors following the same confidence and verification criteria described above. Such a process is repeated until all frames in the sub-map are processed.
Ultimately, this procedure yields a robust set of anchors $\mathcal{A}_k$. Each anchor encapsulates its local 3D coordinate $\mathbf{P}_{local}$ in the sub-map frame and the set of its 2D projection coordinates across the sequence. These anchors constitute the skeleton of the sub-map, preparing for the subsequent non-linear alignment.

\subsection{Bi-directional Anchor Propagation}
\label{subsec:anchor_propagation}

In the CAL$^\text{2}$M framework, adjacent sub-maps share a set of common frames. Since a pixel in the common frame corresponds to the same physical point in the world regardless of which sub-map it belongs to, these frames serve as the bridge for sharing geometric information between sub-maps. We leverage this property to propagate anchors across the sequence, thereby linking different sub-maps.

The propagation mechanism consists of two distinct phases: forward propagation and backward propagation. Forward propagation occurs during the initialization of a new sub-map $\mathcal{S}_k$. Before extracting new anchors, we identify existing anchors from the preceding sub-map $\mathcal{S}_{k-1}$ that are visible in the common frames (the end of $\mathcal{S}_{k-1}$ and the start of $\mathcal{S}_k$). These anchors are directly ``copied'' into $\mathcal{S}_k$. Using their 2D projection coordinates on the common frames and the depth maps estimated within $\mathcal{S}_k$, we back-project them to obtain their new local 3D coordinates $\mathbf{P}_{local}^k$. These propagated anchors serve as the initial seed, after which the standard iterative densification (Sec.~\ref{subsec:anchor_extraction}) is performed to populate the rest of the sub-map. This ensures that historical geometric information is carried forward into the latest estimation.

Conversely, backward propagation is triggered upon the completion of anchor extraction for sub-map $\mathcal{S}_k$. Newly initialized anchors in $\mathcal{S}_k$ may also be visible in the common frames shared with $\mathcal{S}_{k-1}$. We propagate these anchors backward by copying their 2D observations to $\mathcal{S}_{k-1}$. Using the depth information from $\mathcal{S}_{k-1}$, we compute their local 3D coordinates $\mathbf{P}_{local}^{k-1}$ and verify their validity by checking projection consistency against other frames in $\mathcal{S}_{k-1}$. If an anchor is successfully propagated to the start of $\mathcal{S}_{k-1}$, the process recurses, allowing it to be propagated further back to $\mathcal{S}_{k-2}$. This feedback loop allows new geometric discoveries to associate with historical sub-maps. Furthermore, in the event of a loop closure, anchors are propagated between the current sub-map and the historical sub-map via the loop closure frames using the same logic.

Through this bi-directional propagation, we establish a set of ``global anchors''. Each global anchor is associated with a unique physical ID and maintains a list of its local 3D coordinates $\{\mathbf{P}_{i}^a, \mathbf{P}_{j}^a, \dots\}$ across different sub-maps, as well as its 2D projections across different frames. These anchors effectively interconnects the disconnected sub-map skeletons, laying the foundation for the globally consistent mapping.

\subsection{Anchor Fusion and Global Coordinate Estimation}
\label{subsec:anchor_fusion}

Following the propagation stage, we obtain a set of global anchors, where each anchor $a$ is associated with a list of local coordinates $\{\mathbf{P}_{k}^a\}$ across the sub-maps $\{\mathcal{S}_k\}$ in which it is observed. By fusing these local observations, we estimate the anchor's unified global position.

\subsubsection{Weighted Global Fusion}
First, we transform each local coordinate into the global frame using the sub-map's current pose $\mathbf{T}_{w, k}$. The global coordinate $\mathbf{P}_{global}^a$ is computed as the weighted average of these transformed observations,
\begin{equation}
	\mathbf{P}_{global}^a = \frac{\sum_{k \in \mathcal{O}_a} w_{k}^a \cdot (\mathbf{T}_{w, k} \cdot \mathbf{P}_{k}^a)}{\sum_{k \in \mathcal{O}_a} w_{k}^a} ,
\end{equation}
where $\mathcal{O}_a$ is the set of sub-maps observing anchor $a$.
The weight $w_{k}^a$ reflects the reliability of the observation in sub-map $\mathcal{S}_k$. Assuming that points in the central region of the submap offer higher confidence due to richer multi-view constraints, we define the weight to be inversely proportional to the squared Euclidean distance between the anchor position and the optical center $\mathbf{C}_{center}^k$ of the sub-map's central keyframe.
To ensure real-time performance, the fusion is performed incrementally. Specifically, we only update the status of anchors observed in the newly constructed sub-map.



\subsubsection{Topology Preservation via Local Suppression}
While the fusion mechanism integrates multi-view information, a specific challenge arises, namely ``anchor tearing''. Adjacent anchors in the physical world may have different observation counts. This discrepancy in data sources can lead to inconsistent convergence positions, causing the fused anchors to break the original local geometric structure and inducing severe distortions in the final map.


To resolve this, we introduce a local suppression mechanism. Specifically, we prioritize anchors with higher observability. For each anchor $p$, we query its spatial neighbors within a radius $r$ using a KD-Tree. If there exists any neighbor $q$ such that its observation count $N_{obs}(q)$ is strictly greater than $N_{obs}(p)$, we mark anchor $p$ as deactivated,
\begin{equation}
	State(p) = \begin{cases} 
		\text{Deactivated}, & \exists q \in \mathcal{N}_r(p), N_{obs}(q) > N_{obs}(p) \\
		\text{Active}, & \text{otherwise}
	\end{cases} .
	\small
\end{equation}
Deactivated anchors are excluded from constructing the mapping relationship. This ensures that the global skeleton is composed of the most robustly observed points, effectively preventing anchor tearing. The resulting set of active anchors provides a set of consistent control points, defining an accurate mapping from the local sub-map domain to the global domain.

\subsection{Non-Linear Map Deformation}
\label{subsec:tps_deformation}

Having established the mapping relationship via active anchors, the final step is to rectify the dense point cloud using the TPS transform \cite{TPS}. To ensure numerical stability, we explicitly decouple the non-linear deformation from the rigid body transformation. For a sub-map $\mathcal{S}_k$, we map the fused global anchor coordinates back to the local frame using the inverse of the optimized sub-map pose to obtain the target control points $\mathbf{P}'_{target}$. By fitting the TPS model to map the original local anchors to $\mathbf{P}'_{target}$, we isolate the internal non-linear distortion.

We model the non-linear deformation function $\Phi(\mathbf{x})$ for an arbitrary point $\mathbf{x}$ as a combination of an affine transformation and a non-linear warping term based on radial basis functions,
\begin{equation}
	\Phi(\mathbf{x}) = \mathbf{x} \mathbf{A} + \mathbf{t} + \sum_{i=1}^{N} \mathbf{w}_i U(\| \mathbf{x} - \mathbf{p}_i \|) ,
\end{equation}
where $\mathbf{A}$ and $\mathbf{t}$ represent the affine component, $\{\mathbf{p}_i\}$ are the local anchors, and $U(r) = r$ is the radial basis function. The weights $\{\mathbf{w}_i\}$ and affine parameters are solved by minimizing an energy function that balances data fidelity with surface smoothness, governed by a stiffness coefficient $\lambda$. Since the mathematical solution for TPS parameters is a well-established technique \cite{TPS}, we omit the detailed derivation for brevity. The dense point cloud $\mathcal{C}_{raw}$ is finally transformed to the globally consistent state by applying the learned local deformation followed by the global rigid pose: $\mathcal{C}_{final} = \mathbf{T}_{w, k} \cdot \Phi(\mathcal{C}_{raw})$.

\section{Experiments}
\label{sec:experiments}

\subsection{Experimental Setup}
\label{subsec:experimental_setup}

\subsubsection{Datasets}
We evaluate CAL$^\text{2}$M across three diverse datasets to comprehensively assess both localization accuracy and mapping quality under different conditions.

\begin{itemize}
	\item \textbf{KITTI Odometry \cite{KITTI-Odom}:} As one of the most widely adopted benchmarks for long-term outdoor SLAM, we adopt it for evaluating the overall localization performance of compared methods. Following the settings of \cite{VGGT-Long}, we utilize the 11 training sequences (00-10), which contain ground truth poses, to benchmark our system against other typical baselines.
	
	\item \textbf{KITTI-360 \cite{KITTI-360}:}
	 While KITTI Odometry is a standard benchmark, its long trajectories often contain frequent loop closures that can somewhat mask drift accumulation issues. To rigorously test the system's drift-free capability in open-loop scenarios, we employ the KITTI-360 dataset as a supplement. We use 9 sequences in the dataset and processed the first 2,000 frames of each. Crucially, to better assess the long-term stability of compared methods, we disable the loop closure detection module for both our method and all other competitors on this benchmark.
	
	\item \textbf{Argoverse \cite{Argoverse}:} Evaluating mapping accuracy in long-trajectory scenarios is often conflated with localization drift. To isolate and assess the structural quality of different mapping methods, we utilize the Argoverse dataset. Its sequences are relatively shorter, minimizing the impact of cumulative pose errors and allowing for a focused evaluation of reconstruction fidelity. We conduct experiments on the first 10 sequences (0f $\sim$ 44c).
\end{itemize}
%
%

\subsubsection{Implementation Details}
We implemented CAL$^\text{2}$M using PyTorch for the VGFM inference and GTSAM \cite{GTSAM} for the backend PGO. Without specific clarification, VGGT \cite{VGGT} was chosen as the default backbone. All experiments were conducted on a workstation with an RTX6000 GPU. The key hyperparameters are configured as follows:

In the sub-map construction module, the optical flow disparity threshold for keyframe selection is set to $\tau_{flow} = 25$. To balance real-time performance with window size, the maximum number of primary keyframes per sub-map is limited to $N_{max} = 15$, and the number of overlapping common frames between adjacent sub-maps is set to $N_{overlap} = 3$.

In the Primary-Assistant Joint PGO, the local window size $K$ for the odometry constraint is set to 3. The noise model for the prior constraint $E_{prior}$ is set as a $6 \times 6$ diagonal covariance matrix with values of $0.01$. For the odometry constraints ($E_{pri}$ and $E_{ast}$) and loop closure constraints ($E_{loop}$), we assign a diagonal covariance of $0.05$ for the rotational components and $0.1$ for the translational components.

For the test bank construction, a pair of images is considered valid for fundamental matrix estimation if they share at least $N_{feature} = 10$ matched features. A test group is successfully instantiated if it contains at least $N_{pair} = 10$ valid image pairs. The maximum capacity of the test bank is limited to $N_{group} = 5$ groups.

During the anchor extraction, the image is divided into a grid of size $N_{grid} = 24 \times 24$. The relative depth error tolerance factor for anchor projection is set to $\eta_{proj}=2\%$. In the anchor fusion stage, the search radius for local suppression is set to $r = 0.4$. For the non-linear alignment, the stiffness parameter $\lambda$ in the TPS transformation is set to $10^{-4}$.

\subsection{Localization Performance}
\label{subsec:localization_perf}


\subsubsection{KITTI Odometry}
\begin{table*}[t]
	\centering
	\caption{Quantitative Comparison of Absolute Trajectory Error (ATE [m]) on KITTI Odometry Dataset\cite{KITTI-Odom}.}
	\label{tab:kitti_odom}
	\resizebox{\textwidth}{!}{%
		\begin{tabular}{cll|ccccccccccc|cc}
			\toprule
			\textbf{Calib.} & \textbf{Method} & \textbf{Recon.} & \textbf{00} & \textbf{01} & \textbf{02} & \textbf{03} & \textbf{04} & \textbf{05} & \textbf{06} & \textbf{07} & \textbf{08} & \textbf{09} & \textbf{10} & \textbf{Avg} & \textbf{Avg$^*$} \\ \midrule
			- & \textit{Seq. Frames} & - & 4542 & 1101 & 4661 & 801 & 271 & 2761 & 1101 & 1101 & 4071 & 1591 & 1201 & 2109 & 2210 \\
			- & \textit{Seq. Length (m)} & - & 3724 & 2453 & 5067 & 561 & 394 & 2206 & 1233 & 650 & 3223 & 1705 & 920 & 2012.24 & 1968.15 \\
			- & \textit{Contains Loop} & - & \checkmark & $\times$ & \checkmark & $\times$ & $\times$ & \checkmark & \checkmark & \checkmark & $\times$ & \checkmark & $\times$ & - & - \\ \midrule
			
			\multirow{5}{*}{$\checkmark$} 
			& ORB-SLAM2 \cite{ORB-SLAM2} & Sparse & \textbf{6.03} & 508.34 & \textbf{14.76} & \textbf{1.02} & 1.57 & \textbf{4.04} & \textbf{11.16} & \underline{2.19} & \textbf{38.85} & \textbf{8.39} & \textbf{6.63} & 54.82 & \textbf{9.46} \\
			& LDSO \cite{LDSO} & Sparse & 9.32 & \underline{11.68} & \underline{31.98} & 2.85 & 1.22 & \underline{5.10} & 13.55 & 2.96 & 129.02 & \underline{21.64} & 17.36 & \textbf{22.43} & \underline{23.50} \\
			& DPV-SLAM \cite{DPV-SLAM} & Sparse & 112.80 & \textbf{11.50} & 123.53 & \underline{2.50} & \underline{0.81} & 57.80 & 54.86 & 18.77 & \underline{110.49} & 76.66 & \underline{13.65} & 53.03 & 57.19 \\
			& DPV-SLAM++ \cite{DPV-SLAM} & Sparse & \underline{8.30} & 11.86 & 39.64 & \underline{2.50} & \textbf{0.78} & 5.74 & \underline{11.60} & \textbf{1.52} & 110.90 & 76.70 & 13.70 & \underline{25.75} & 27.14 \\
			& DROID-SLAM \cite{DROID-SLAM} & Dense & 92.10 & 344.60 & 107.61 & 2.38 & 1.00 & 118.50 & 62.47 & 21.78 & 161.60 & 72.32 & 118.70 & 100.28 & 75.85 \\ 
			 \midrule
			
			\multirow{6}{*}{$\times$} 
			& MASt3R-SLAM \cite{Mast3rSLAM} & Dense & TL & TL & TL & TL & TL & TL & TL & TL & TL & TL & TL & / & / \\
			& CUT3R \cite{CUT3R} & Dense & OOM & OOM & OOM & 148.07 & 22.31 & OOM & OOM & OOM & OOM & OOM & OOM & / & / \\
			& VGGT-Long \cite{VGGT-Long} & Dense & \underline{8.67} & \underline{121.17} & \underline{32.08} & \underline{6.12} & 4.23 & \underline{8.31} & \underline{5.34} & \underline{4.63} & \underline{53.10} & \underline{41.99} & \underline{18.37} & \underline{27.63} & \underline{18.28} \\
			& VGGT-SLAM \cite{VGGT-SLAM} & Dense & FAIL & 309.05 & FAIL & 27.17 & \underline{2.91} & FAIL & FAIL & FAIL & 269.03 & FAIL & 223.04 & / & / \\
			& \textbf{CAL$^\text{2}$M} & Dense & \textbf{4.36} & \textbf{72.30} & \textbf{22.81} & \textbf{3.29} & \textbf{1.35} & \textbf{6.21} & \textbf{3.88} & \textbf{1.52} & \textbf{21.71} & \textbf{11.85} & \textbf{12.55} & \textbf{14.71} & \textbf{8.95} \\ \bottomrule
			\multicolumn{16}{l}{\footnotesize \textbf{Bold}: Best result; \underline{Underline}: Second best result within each category (Calibrated / Calibration-Free). } \\
			\multicolumn{16}{l}{\footnotesize TL: Tracking Lost; OOM: Out Of Memory; FAIL: Numerical Instability/Crash. Avg$^*$: Average ATE excluding Sequence 01 (Highway scenario).}
		\end{tabular}%
	}
\end{table*}

We first conducted a comprehensive evaluation on the 11 training sequences (00-10) of the KITTI Odometry dataset \cite{KITTI-Odom}. The baselines include two categories: calibrated methods, which include traditional geometry-based systems \cite{ORB-SLAM2, LDSO} and learning-based approaches \cite{DROID-SLAM, DPV-SLAM} that rely on accurate camera calibration; and calibration-free methods, covering recent VGFM-based approaches \cite{Mast3rSLAM, CUT3R, VGGT-Long, VGGT-SLAM}.
As reported in Table~\ref{tab:kitti_odom}, CAL$^\text{2}$M achieves the best overall performance across the 11 sequences compared with other calibration-free methods. 
Furthermore, it delivers highly competitive results even against established calibrated systems like \cite{LDSO} and \cite{ORB-SLAM2}. These results validate the effectiveness of CAL$^\text{2}$M in rectifying scale drift, proving its capability for robust long-term localization without pre-calibration.

\begin{table*}[t]
	\centering
	\caption{Quantitative Comparison of ATE [m] on KITTI-360 dataset \cite{KITTI-360} (Loop Closure Disabled).}
	\label{tab:kitti_360}
	\resizebox{\textwidth}{!}{%
		\begin{tabular}{ll|ccccccccc|cc}
			\toprule
			\textbf{Method} & \textbf{Calib.} & \textbf{00} & \textbf{02} & \textbf{03} & \textbf{04} & \textbf{05} & \textbf{06} & \textbf{07} & \textbf{09} & \textbf{10} & \textbf{Avg} & \textbf{Avg$^*$} \\ \midrule
			\textit{Seq. Frames} & - & 2000 & 2000 & 1030 & 2000 & 2000 & 2000 & 2000 & 2000 & 2000 & 1892 & 1879 \\ 
			\textit{Seq. Length (m)} & - & 1353 & 1528 & 1379 & 1891 & 1444 & 1429 & 4058 & 1952 & 2245 & 1919.89 & 1652.63 \\ \midrule
			DROID-SLAM \cite{DROID-SLAM} & $\checkmark$ & \underline{16.46} & \underline{16.96} & \textbf{5.18} & \textbf{9.99} & \underline{19.11} & \underline{14.23} & 436.94 & \underline{23.68} & \underline{46.05} & \underline{65.40} & \underline{18.96} \\ \midrule
			MASt3R-SLAM \cite{Mast3rSLAM} & $\times$ & TL & TL & TL & TL & TL & TL & TL & TL & TL & / & / \\
			CUT3R \cite{CUT3R} & $\times$ & OOM & OOM & OOM & OOM & OOM & OOM & OOM & OOM & OOM & / & / \\
			VGGT-Long \cite{VGGT-Long} & $\times$ & 50.92 & 38.90 & 21.07 & 155.41 & 39.81 & 35.30 & \underline{331.46} & 113.73 & 161.74 & 105.37 & 77.11 \\
			VGGT-SLAM \cite{VGGT-SLAM} & $\times$ & 120.99 & 194.44 & 336.56 & 198.26 & 139.71 & 187.22 & 789.07 & 325.87 & 494.29 & 309.60 & 249.67 \\
			\textbf{CAL$^2$M} & $\times$ & \textbf{10.83} & \textbf{9.35} & \underline{12.60} & \underline{17.93} & \textbf{9.95} & \textbf{9.59} & \textbf{64.81} & \textbf{14.36} & \textbf{22.25} & \textbf{19.07} & \textbf{13.36} \\ \bottomrule
			\multicolumn{13}{l}{\footnotesize TL: Tracking failure; OOM: Out Of Memory. Avg$^*$: Average ATE excluding Sequence 07 (Highway scenario).}
		\end{tabular}%
	}
\end{table*}

\subsubsection{KITTI-360}
To strictly evaluate the localization stability over kilometer-scale trajectories, we tested on 9 sequences from the KITTI-360 dataset \cite{KITTI-360}. In this experiment, loop closure detection was disabled for all methods. The comparison includes dense-mapping capable approaches: the calibrated DROID-SLAM \cite{DROID-SLAM} and the calibration-free MASt3R-SLAM \cite{Mast3rSLAM}, CUT3R \cite{CUT3R}, VGGT-Long \cite{VGGT-Long}, and VGGT-SLAM \cite{VGGT-SLAM}.
The quantitative results are summarized in Table~\ref{tab:kitti_360}. CAL$^\text{2}$M exhibits superior stability, outperforming all other competitors.
Such a superior performance is attributed to our framework's ability to effectively eliminate scale drift via the assistant eye, while simultaneously mitigating the cumulative errors caused by intrinsic fluctuations through the epipolar-guided pose correction model.

%
%

\subsubsection{Synchronization Analysis}

A key feature of CAL$^\text{2}$M is its independence from hardware clock synchronization. To validate this, we simulated an asynchronous setup on both KITTI Odometry \cite{KITTI-Odom} and KITTI-360 \cite{KITTI-360} by feeding odd-indexed frames to the primary stream and even-indexed frames to the assistant one.
The comparison between synchronous and asynchronous modes is presented in Table~\ref{tab:sync_async_combined}. The results indicate that our algorithm maintains consistent performance on both datasets regardless of the synchronization setup. 
Interestingly, in certain sequences of KITTI Odometry \cite{KITTI-Odom}, the asynchronous mode yields slightly better accuracy. We hypothesize this is because the alternating frame selection creates more uniform spatial distributions of observations, increasing the performance of VGFM. 
This experiment validates that CAL$^\text{2}$M offers high freedom in sensor configuration, requiring neither spatial calibration nor temporal synchronization.

\begin{table}[h]
	\centering
	\caption{Localization Accuracy (ATE [m]) under Different Sync. Mode.}
	\label{tab:sync_async_combined}
	\resizebox{\columnwidth}{!}{%
		\begin{tabular}{c|l|cccccc}
			\toprule
			\textbf{Dataset} & \textbf{Mode} & \textbf{00} & \textbf{01} & \textbf{02} & \textbf{03} & \textbf{04} & \textbf{05} \\ \midrule
			\multirow{5}{*}{\rotatebox{90}{KITTI-Odom}} 
			& Sync. & 4.36 & \textbf{72.30} & 22.81 & 3.29 & 1.35 & 6.21 \\
			& Async. & \textbf{4.02} & 73.99 & \textbf{21.34} & \textbf{2.87} & \textbf{0.70} & \textbf{4.91} \\ \cmidrule{2-8}
			& \textbf{Seq.} & \textbf{06} & \textbf{07} & \textbf{08} & \textbf{09} & \textbf{10} & \textbf{Avg.} \\ \cmidrule{2-8}
			& Sync. & \textbf{3.88} & \textbf{1.52} & 21.71 & \textbf{11.85} & 12.55 & 14.71 \\
			& Async. & 4.61 & 1.75 & \textbf{21.42} & 13.27 & \textbf{10.92} & \textbf{14.53} \\ 
			\midrule[\heavyrulewidth] 
			\textbf{Dataset} & \textbf{Mode} & \textbf{00} & \textbf{02} & \textbf{03} & \textbf{04} & \textbf{05} & \textbf{-} \\ \midrule
			\multirow{5}{*}{\rotatebox{90}{KITTI-360}}
			& Sync. & \textbf{10.83} & \textbf{9.35} & \textbf{12.60} & \textbf{17.93} & \textbf{9.95} & - \\
			& Async. & 12.81 & 10.06 & 13.66 & 18.05 & 12.73 & - \\ \cmidrule{2-8}
			& \textbf{Seq.} & \textbf{06} & \textbf{07} & \textbf{09} & \textbf{10} & \textbf{Avg.} & \textbf{-} \\ \cmidrule{2-8}
			& Sync. & \textbf{9.59} & 64.81 & 14.36 & \textbf{22.25} & \textbf{19.07} & - \\
			& Async. & 13.12 & \textbf{55.89} & \textbf{14.26} & 28.82 & 19.93 & - \\ \bottomrule
		\end{tabular}%
	}
\end{table}


\begin{figure*}[t]
	\centering
	\includegraphics[width=2.0\columnwidth]{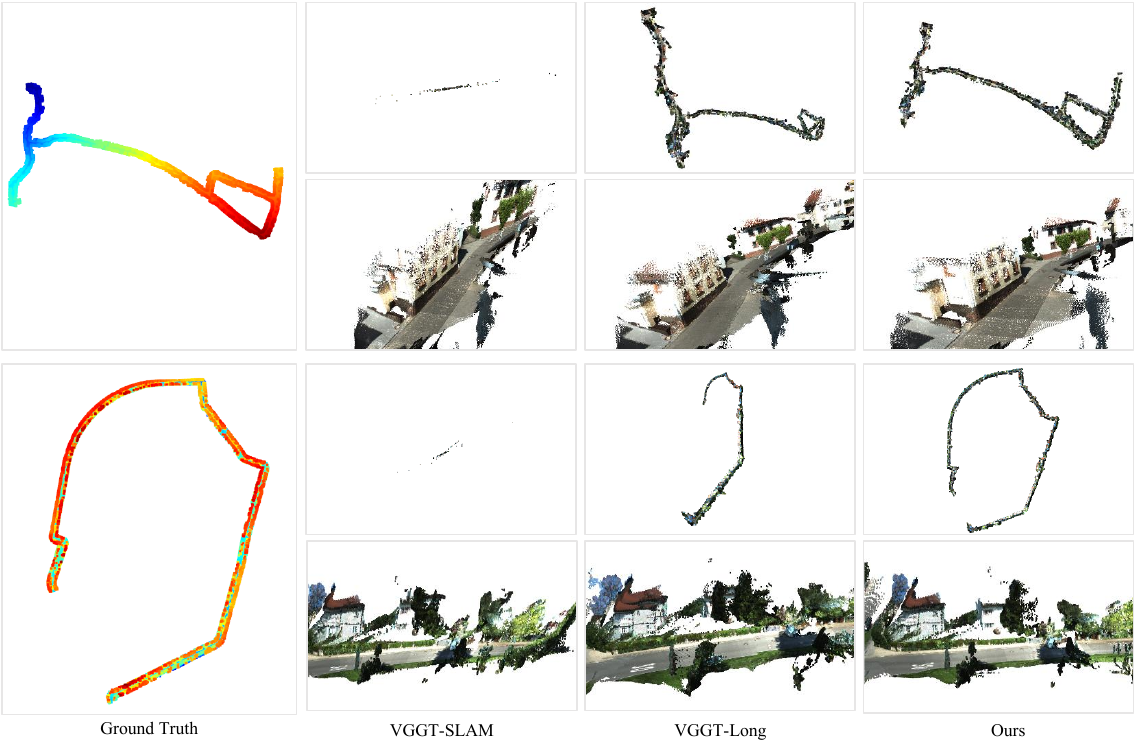}
	\vspace{0.5mm}
	\caption{Qualitative comparison of 3D reconstruction results on KITTI-360 \cite{KITTI-360}. The figure displays Sequence 06 (top) and Sequence 09 (bottom). For each sequence, the columns correspond to (from left to right): Ground Truth (LiDAR), VGGT-SLAM \cite{VGGT-SLAM}, VGGT-Long \cite{VGGT-Long}, and our CAL$^\text{2}$M. Within each sub-figure, the upper view illustrates the global point cloud, while the lower view provides a zoomed-in perspective of local structural details.}
	\label{fig:kitti360_map}
\end{figure*}

\subsection{Mapping Performance}
\label{subsec:mapping_perf}


\subsubsection{Qualitative Evaluation on KITTI-360}
Since the overall accuracy of long-term mapping is heavily dependent on localization accuracy, on KITTI-360 \cite{KITTI-360}, we mainly conduct the qualitative evaluation.
We compare the reconstruction quality of CAL$^\text{2}$M against two typical VGFM-based approaches: VGGT-Long \cite{VGGT-Long} and VGGT-SLAM \cite{VGGT-SLAM}. To serve as a reference, we also generate ground truth maps using LiDAR point clouds accumulated with ground truth poses. Visual comparisons are given in Fig.~\ref{fig:kitti360_map}.

From a global perspective, the Sim3-based alignment in \cite{VGGT-Long} suffers from severe scale drift, while the SL4-based alignment in \cite{VGGT-SLAM} leads to accumulated structural distortions over long trajectories. In contrast, CAL$^\text{2}$M maintains global stability, producing reconstruction results that align closely with the ground truth map geometry.

In terms of local details, VGGT-Long \cite{VGGT-Long} display noticeable geometric misalignments between adjacent sub-maps, and VGGT-SLAM \cite{VGGT-SLAM} introduces obvious non-physical distortions in local structures caused by the instability of the SL4 transformation. Leveraging our global anchor propagation and non-linear TPS alignment, CAL$^\text{2}$M preserves excellent local detail fidelity while achieving seamless transitions between sub-maps, effectively eliminating geometric misalignment.

\subsubsection{Evaluation on Argoverse}

\begin{table*}[t]
	\centering
	\caption{Quantitative Evaluation of Dense Mapping Quality on Argoverse \cite{Argoverse}.}
	\label{tab:argoverse_map}
	\resizebox{0.95\textwidth}{!}{%
		\begin{tabular}{ll|cccccccccc|c}
			\toprule
			\textbf{Method} & \textbf{Metric [m]} & \textbf{00} & \textbf{01} & \textbf{02} & \textbf{03} & \textbf{04} & \textbf{05} & \textbf{06} & \textbf{07} & \textbf{08} & \textbf{09} & \textbf{Avg.} \\ \midrule
			
			\multirow{3}{*}{\shortstack[l]{DROID-SLAM \cite{DROID-SLAM} }} 
			& Chamfer $\downarrow$ & 5.35 & 2.29 & 2.04 & 2.31 & 4.61 & 2.05 & 2.18 & 2.16 & \underline{1.73} & 3.48 & 2.82 \\
			& Accuracy $\downarrow$ & 3.30 & 1.94 & 2.52 & 2.39 & 4.20 & 1.85 & 2.10 & 2.86 & 1.68 & 3.60 & 2.65 \\
			& Completeness $\downarrow$ & 7.41 & 2.65 & \textbf{1.55} & 2.23 & 5.03 & 2.25 & 2.26 & \textbf{1.46} & \textbf{1.77} & 3.36 & 3.00 \\ \midrule
			
			\multirow{3}{*}{\shortstack[l]{MASt3R-SLAM \cite{Mast3rSLAM}}} 
			& Chamfer $\downarrow$ & 2.88 & 3.53 & 2.99 & 3.88 & 4.17 & 3.23 & 3.58 & 4.19 & 3.21 & 2.54 & 3.42 \\
			& Accuracy $\downarrow$ & 3.23 & 4.40 & 3.47 & 5.00 & 5.08 & 3.91 & 4.28 & 3.69 & 3.95 & 2.75 & 3.98 \\
			& Completeness $\downarrow$ & 2.52 & 2.66 & 2.52 & 2.77 & 3.26 & 2.55 & 2.87 & 4.69 & 2.46 & 2.33 & 2.86 \\ \midrule
			
			\multirow{3}{*}{\shortstack[l]{CUT3R \cite{CUT3R}}} 
			& Chamfer $\downarrow$ & 1.77 & \underline{1.77} & 2.23 & \underline{1.52} & 2.10 & 2.26 & 1.81 & 1.74 & \underline{1.73} & 1.81 & {1.87} \\
			& Accuracy $\downarrow$ & 1.22 & \underline{1.31} & 1.68 & \textbf{0.93} & 1.50 & 1.68 & \underline{1.01} & 1.09 & \underline{1.07} & 1.13 & 1.26 \\
			& Completeness $\downarrow$ & 2.32 & \underline{2.23} & 2.77 & 2.10 & 2.70 & 2.84 & 2.62 & 2.38 & 2.39 & 2.48 & 2.48 \\ \midrule
			
			\multirow{3}{*}{\shortstack[l]{VGGT-Long \cite{VGGT-Long}}} 
			& Chamfer $\downarrow$ & 1.67 & 2.22 & {1.34} & 1.55 & \underline{1.76} & \underline{1.48} & 2.00 & \underline{1.57} & 2.09 & 2.30 & \underline{1.80} \\
			& Accuracy $\downarrow$ & 1.02 & 1.49 & 0.93 & \underline{1.11} & \underline{1.38} & \textbf{0.65} & 1.17 & \underline{1.07} & 1.37 & 2.07 & \underline{1.23} \\
			& Completeness $\downarrow$ & 2.32 & 2.96 & 1.74 & \underline{1.98} & \underline{2.15} & \underline{2.30} & 2.82 & 2.07 & 2.80 & 2.53 & 2.37 \\ \midrule
			
			\multirow{3}{*}{\shortstack[l]{VGGT-SLAM \cite{VGGT-SLAM}}} 
			& Chamfer $\downarrow$ & \underline{1.54} & 2.35 & \underline{1.31} & 2.23 & 2.65 & 1.79 & \underline{1.61} & 1.78 & 2.23 & \underline{1.28} & 1.88 \\
			& Accuracy $\downarrow$ & \underline{0.86} & 2.53 & \underline{0.84} & 1.99 & 2.97 & {0.75} & 1.19 & 1.25 & 1.63 & \underline{0.99} & 1.50 \\
			& Completeness $\downarrow$ & \underline{2.22} & \textbf{2.17} & 1.79 & 2.48 & 2.33 & 2.83 & \underline{2.03} & 2.30 & 2.83 & \underline{1.58} & \underline{2.26} \\ \midrule
			
			\multirow{3}{*}{\textbf{CAL$^\text{2}$M}} 
			& Chamfer $\downarrow$ & \textbf{1.45} & \textbf{1.67} & \textbf{1.22} & \textbf{1.38} & \textbf{1.66} & \textbf{1.31} & \textbf{1.30} & \textbf{1.39} & \textbf{1.72} & \textbf{1.05} & \textbf{1.41} \\
			& Accuracy $\downarrow$ & \textbf{0.81} & \textbf{1.08} & \textbf{0.78} & \textbf{0.93} & \textbf{1.19} & \underline{0.73} & \textbf{0.84} & \textbf{1.03} & \textbf{1.06} & \textbf{0.73} & \textbf{0.92} \\
			& Completeness $\downarrow$ & \textbf{2.08} & 2.26 & \underline{1.66} & \textbf{1.82} & \textbf{2.13} & \textbf{1.89} & \textbf{1.75} & \underline{1.74} & \underline{2.38} & \textbf{1.37} & \textbf{1.91} \\ \bottomrule
			
		\end{tabular}%
	}
\end{table*}

\begin{figure*}[p]
	\centering
	\subfloat[Seq. 01]{
		\includegraphics[width=0.9\textwidth]{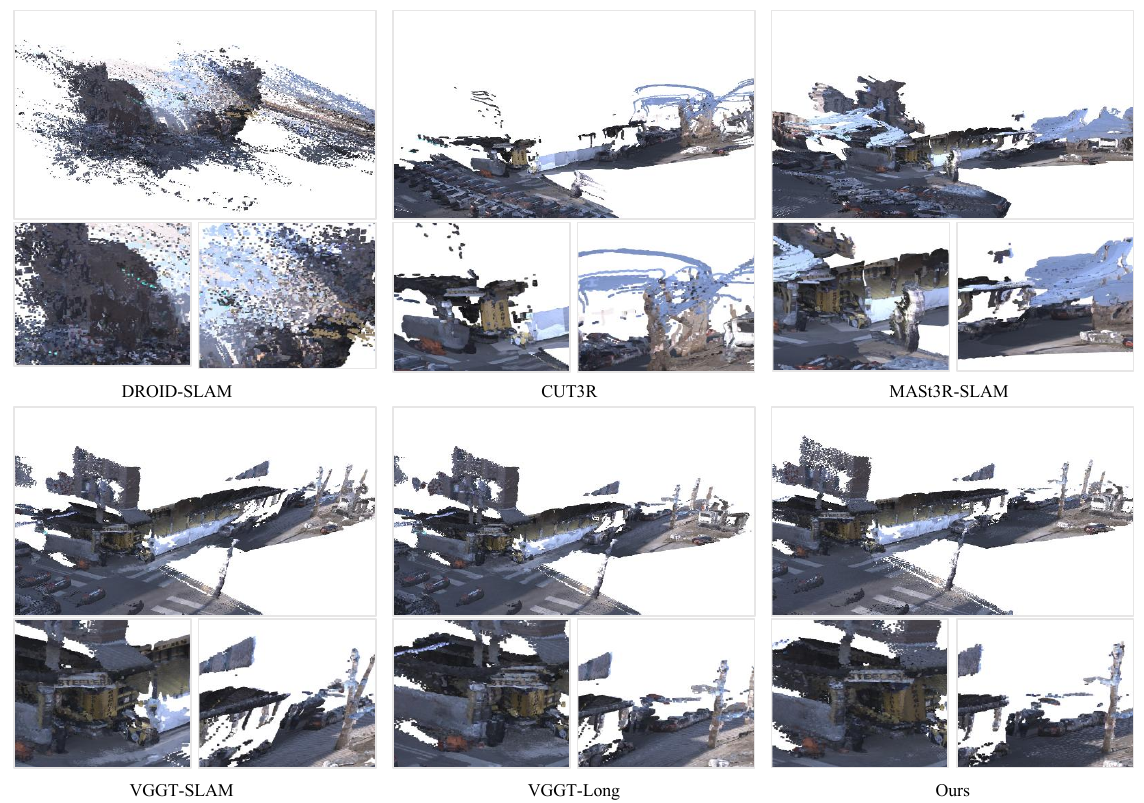}
	}
	\\
	\subfloat[Seq. 04]{
		\includegraphics[width=0.9\textwidth]{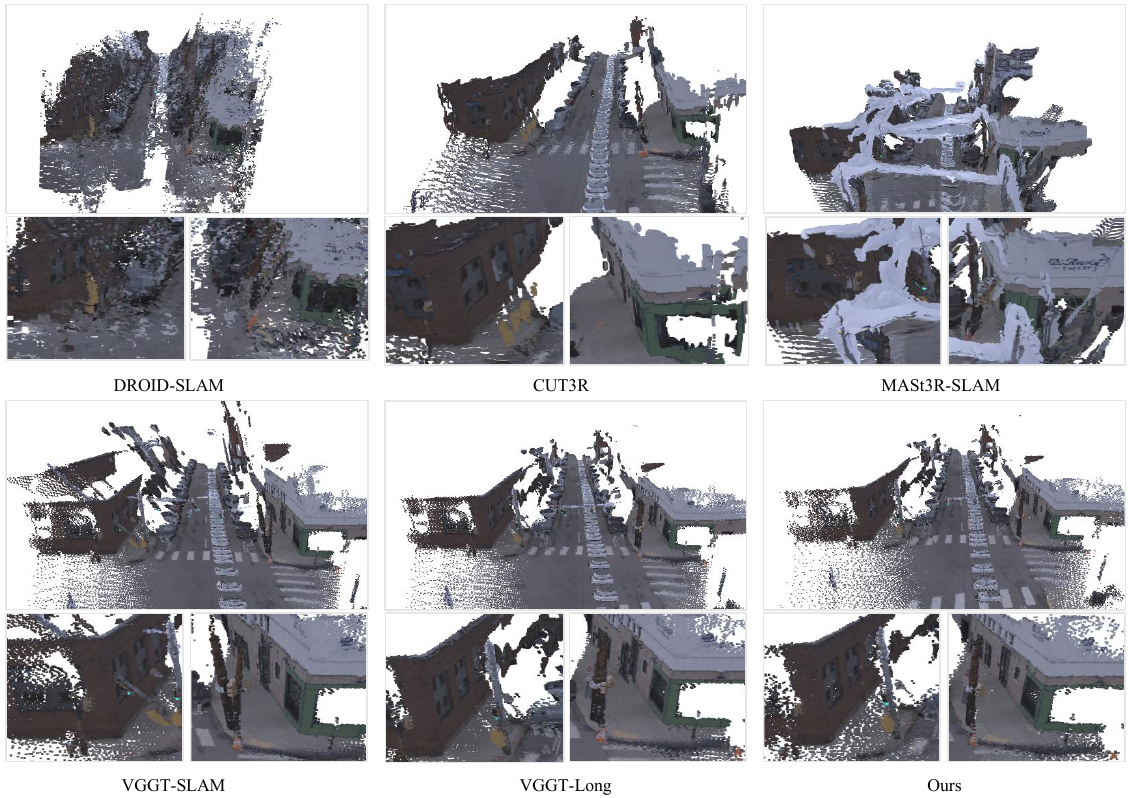}
	}
	\caption{Qualitative comparison of dense reconstruction results on Argoverse \cite{Argoverse} Sequence 01 (top) and Sequence 04 (bottom). The figure displays results from six methods: DROID-SLAM \cite{DROID-SLAM}, CUT3R \cite{CUT3R}, MASt3R-SLAM \cite{Mast3rSLAM}, VGGT-SLAM \cite{VGGT-SLAM}, VGGT-Long \cite{VGGT-Long}, and our CAL$^\text{2}$M. }
	\label{fig:argoverse_map}
\end{figure*}

We conducted a comprehensive evaluation on the Argoverse dataset, comparing CAL$^\text{2}$M against all dense-mapping capable methods \cite{DROID-SLAM, Mast3rSLAM, CUT3R,  VGGT-Long, VGGT-SLAM}. The accumulated LiDAR point clouds are used as the ground truth.

For quantitative assessment, we employ three standard metrics: Accuracy, Completeness, and Chamfer Distance (the average of accuracy and completeness). As shown in Table~\ref{tab:argoverse_map}, our method achieves the best performance across almost all sequences, which corroborates the high accuracy and stability of our mapping strategy.

Qualitatively,  as shown in Fig. \ref{fig:argoverse_map}, CAL$^\text{2}$M and VGGT-Long \cite{VGGT-Long} produce the visually cleanest structures among the compared methods, avoiding the severe noise and warping observed in other approaches. However, a critical distinction lies in consistency. While VGGT-Long \cite{VGGT-Long} exhibits visible geometric misalignments at sub-map boundaries, our method achieves a coherent global structure. By effectively rectifying internal non-linear distortions via anchors, CAL$^\text{2}$M ensures that the map is not only locally clean but also globally consistent, yielding the highest quality reconstruction.

\subsection{Generalization Ability on Different Backbones}
\label{subsec:generalization}

A core advantage of CAL$^\text{2}$M is its plug-and-play architecture, allowing it to integrate seamlessly with arbitrary VGFMs. To evaluate such generalization capability, we extended our experiments beyond the default VGGT \cite{VGGT} backbone to include two other mainstream models: Pi3 \cite{Pi3} and MapAnything \cite{MapAnything}. For each backbone, we compared our framework against two baseline strategies: the Sim3-based alignment (VGGT-Long \cite{VGGT-Long}) and the SL4-based alignment (VGGT-SLAM \cite{VGGT-SLAM}).

\begin{table*}[t]
	\centering
	\caption{Generalization of Localization Performance (ATE [m]) on KITTI-360 \cite{KITTI-360} across Different Backbones.}
	\label{tab:backbone_loc}
	\resizebox{0.95\textwidth}{!}{%
		\begin{tabular}{ll|ccccccccc|c}
			\toprule
			\textbf{Backbone} & \textbf{Alignment} & \textbf{00} & \textbf{02} & \textbf{03} & \textbf{04} & \textbf{05} & \textbf{06} & \textbf{07} & \textbf{09} & \textbf{10} & \textbf{Avg.} \\ \midrule
			
			\multirow{3}{*}{\textbf{VGGT} \cite{VGGT}} 
			& Sim3 & \underline{71.33} & \underline{98.71} & \underline{104.24} & \underline{195.00} & \underline{91.63} & \underline{72.52} & \underline{718.59} & \underline{222.41} & \underline{245.41} & \underline{202.20} \\
			& SL4 & 155.33 & 194.82 & 360.72 & 200.23 & 140.95 & 200.20 & 1048.41 & 324.93 & 503.42 & 347.67 \\
			& \textbf{Ours} & \textbf{10.83} & \textbf{9.35} & \textbf{12.60} & \textbf{17.93} & \textbf{9.95} & \textbf{9.59} & \textbf{64.81} & \textbf{14.36} & \textbf{22.25} & \textbf{19.07} \\ \midrule
			
			\multirow{3}{*}{\textbf{Pi3} \cite{Pi3}} 
			& Sim3 & \underline{63.99} & \underline{27.84} & \underline{9.73} & \underline{107.36} & \underline{29.84} & \underline{30.52} & \textbf{50.86} & \underline{121.59} & \underline{33.09} & \underline{52.76} \\
			& SL4 & 117.40 & 193.19 & 288.03 & 200.44 & 140.92 & 200.56 & 1051.39 & 325.11 & 503.94 & 335.66 \\
			& \textbf{Ours} & \textbf{11.37} & \textbf{11.05} & \textbf{9.34} & \textbf{29.72} & \textbf{11.66} & \textbf{7.07} & \underline{53.42} & \textbf{17.79} & \textbf{26.62} & \textbf{19.78} \\ \midrule
			
			\multirow{3}{*}{\textbf{MapAnything} \cite{MapAnything}} 
			& Sim3 & \underline{73.17} & \underline{39.94} & \underline{71.57} & \underline{182.06} & \underline{71.61} & \underline{30.50} & \underline{392.56} & \underline{175.06} & \underline{128.62} & \underline{129.45} \\
			& SL4 & 155.39 & 194.00 & 363.83 & 199.14 & 140.85 & 200.36 & 1052.17 & 325.09 & 485.05 & 346.21 \\
			& \textbf{Ours} & \textbf{11.48} & \textbf{7.77} & \textbf{40.38} & \textbf{14.40} & \textbf{17.17} & \textbf{9.03} & \textbf{124.45} & \textbf{13.05} & \textbf{11.62} & \textbf{27.71} \\ \bottomrule
			
		\end{tabular}%
	}
\end{table*}
\subsubsection{Localization Analysis on KITTI-360}
We first assessed the localization performance of these combinations on the KITTI-360 dataset \cite{KITTI-360}. As presented in Table~\ref{tab:backbone_loc}, CAL$^\text{2}$M consistently delivers stable and superior results across all three backbones, significantly outperforming the Sim3 and SL4 baselines in every configuration.

It is worth noting that while the combinations with VGGT \cite{VGGT} and Pi3 \cite{Pi3} exhibit high stability, the results using MapAnything \cite{MapAnything} show slightly higher variance. We attribute this to MapAnything's design philosophy, which heavily prioritizes mapping quality. Consequently, in highly dynamic regions, MapAnything tends to overfit to the changing environment, occasionally compromising the accurate localization. Nevertheless, our framework still effectively mitigates these errors compared to linear alignment baselines.

\begin{table}[h]
	\centering
	\caption{Generalization of Mapping Quality [m] across Different Backbones on Argoverse \cite{Argoverse}.}
	\label{tab:backbone_map}
	\resizebox{\columnwidth}{!}{%
		\begin{tabular}{llccc}
			\toprule
			\textbf{Backbone} & \textbf{Alignment} & \textbf{Accuracy} $\downarrow$ & \textbf{Completeness} $\downarrow$ & \textbf{Chamfer} $\downarrow$ \\ \midrule
			
			\multirow{3}{*}{\textbf{VGGT} \cite{VGGT}} 
			& Sim3 & \underline{1.23} & 2.37 & \underline{1.80} \\
			& SL4 & 1.50 & \underline{2.26} & 1.88 \\
			& \textbf{Ours} & \textbf{0.92} & \textbf{1.91} & \textbf{1.41} \\ \midrule
			
			\multirow{3}{*}{\textbf{Pi3} \cite{Pi3}} 
			& Sim3 & \underline{1.68} & \underline{2.18} & \underline{1.93} \\
			& SL4 & 2.05 & 2.25 & 2.15 \\
			& \textbf{Ours} & \textbf{1.43} & \textbf{1.82} & \textbf{1.63} \\ \midrule
			
			\multirow{3}{*}{\textbf{MapAnything} \cite{MapAnything}} 
			& Sim3 & 2.64 & \underline{2.88} & \underline{2.76} \\
			& SL4 & \underline{2.57} & 3.51 & 3.04 \\
			& \textbf{Ours} & \textbf{1.48} & \textbf{2.23} & \textbf{1.85} \\ \bottomrule
			
		\end{tabular}%
	}
\end{table}

\subsubsection{Mapping Analysis on Argoverse}
We further evaluated the mapping quality on the Argoverse dataset \cite{Argoverse} to assess the generalization capability of our framework across different backbone models. The quantitative results, summarized in Table~\ref{tab:backbone_map}, demonstrate that CAL$^\text{2}$M consistently yields the best performance across all evaluated backbones \cite{VGGT, Pi3, MapAnything}. Regardless of the specific characteristics of the underlying model, our approach significantly outperforms both the Sim3-based and SL4-based alignment baselines in terms of Accuracy, Completeness, and Chamfer Distance. This universal superiority confirms that our proposed strategies—ranging from the assistant-eye constraints to the non-linear anchor alignment—are backbone-agnostic and possess strong generalization capabilities, effectively enhancing the geometric accuracy of arbitrary visual geometry foundation models without being tied to a specific architecture.

\begin{table}[h]
	\centering
	\caption{Quantitative Evaluation of Intrinsic Parameter Stability (Average Focal Length Error [px]).}
	\label{tab:intrinsic_error}
	\resizebox{\columnwidth}{!}{%
		\begin{tabular}{lcc}
			\toprule
			\textbf{Method} & \textbf{KITTI-Odom \cite{KITTI-Odom}} & \textbf{KITTI-360} \cite{KITTI-360} \\ \midrule
			Raw VGFM Output & 34.67 & 46.24 \\
			Naive Mean Correction & 25.50 & 46.21 \\
			\textbf{Ours (Intrinsic Search)} & \textbf{19.89} & \textbf{11.56} \\ \midrule
			\textbf{Success Rate} & 81.8\% (9/11) & 88.9\% (8/9) \\ \bottomrule
		\end{tabular}%
	}
	\vspace{-3mm}
\end{table}

\begin{figure*}[t]
	\centering
	\subfloat{
		\includegraphics[width=0.48\textwidth]{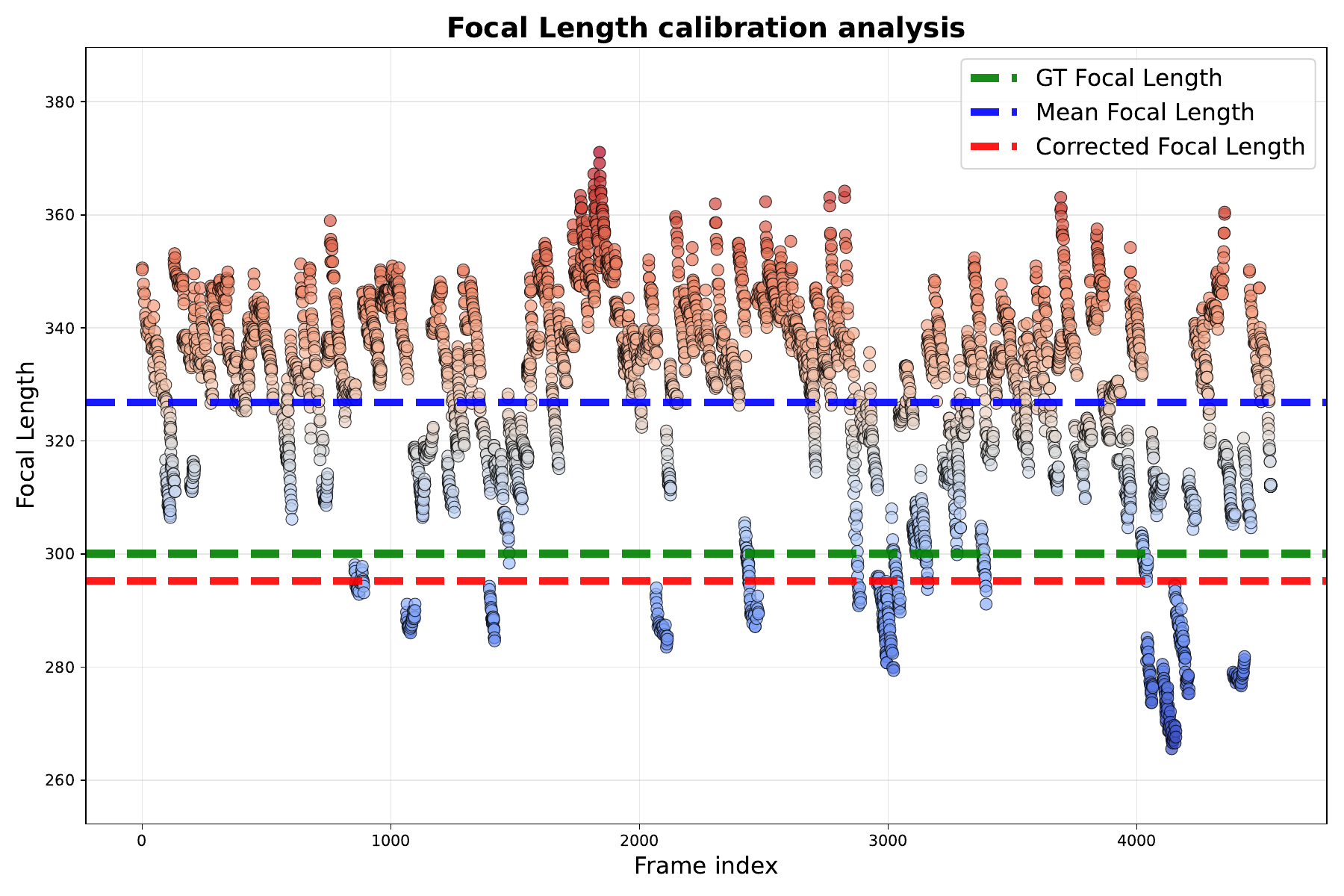}
	}
	\hfill
	\subfloat{
		\includegraphics[width=0.48\textwidth]{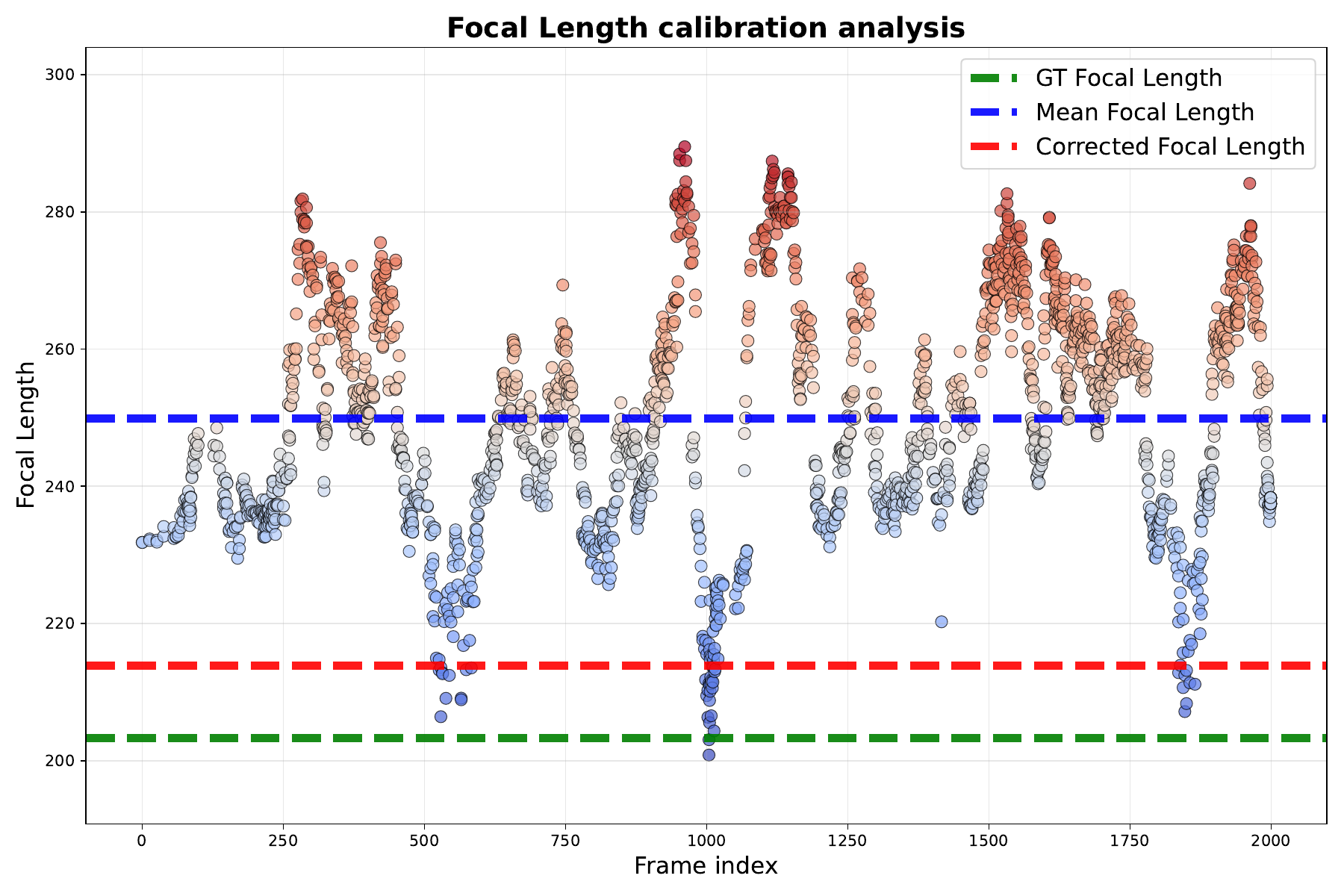}
	}
	\caption{Qualitative evaluation of intrinsic estimation stability. The plots display the focal length estimation over time for KITTI-Odom \cite{KITTI-Odom} Seq. 00 (left) and KITTI-360 \cite{KITTI-360} Seq. 00 (right). The scatter points represent the fluctuating raw outputs from the VGFM. The \textcolor{green}{green dashed line} indicates the Ground Truth, the \textcolor{blue}{blue dashed line} indicates the average focal length of all estimations, and the \textcolor{red}{red line} shows the corrected value determined by online intrinsic search.}
	\label{fig:intrinsic_curves}
\end{figure*}

\subsection{Stability Evaluation}
\label{subsec:stability}

\subsubsection{Intrinsic Estimation Stability}
The intrinsics estimated by VGFMs often exhibit significant fluctuations across different sub-maps. To validate the effectiveness of our online intrinsic search module, we compare its performance against a ``Naive Mean'' strategy, which simply averages all raw VGFM outputs.
The quantitative results on KITTI Odometry \cite{KITTI-Odom} and KITTI-360 \cite{KITTI-360} are presented in Table~\ref{tab:intrinsic_error}. Our method yields lower average errors compared to both the raw output and the naive mean strategy. Qualitative plots in Fig.~\ref{fig:intrinsic_curves} further visualize this improvement, showing that our corrected intrinsics converge closely to the ground truth, whereas the raw predictions fluctuate wildly.

Regarding success rate, our module successfully constructed valid test banks and performed correction in 17 out of the 20 tested sequences. The three exceptions were KITTI-Odom \cite{KITTI-Odom} Seq. 03, Seq. 04 (due to insufficient trajectory length) and KITTI-360 \cite{KITTI-360} Seq. 07 (a highway scenario with sparse features). This high success rate confirms the robustness of our intrinsic search algorithm in diverse environments.

\begin{table}[h]
	\centering
	\caption{Scale Consistency Performance on KITTI-360 \cite{KITTI-360}.}
	\label{tab:scale_drift}
	\resizebox{0.9\columnwidth}{!}{%
		\begin{tabular}{l|ccccc}
			\toprule
			\textbf{Seq.} & \textbf{00} & \textbf{02} & \textbf{03} & \textbf{04} & \textbf{05} \\ \midrule
			VGGT-L \cite{VGGT-Long} (Err.) & 1.157 & 1.263 & 0.642 & 1.641 & 0.832 \\
			VGGT-L \cite{VGGT-Long} (Std.) & 0.849 & 0.822 & 0.311 & 2.061 & 0.617 \\ \midrule
			\textbf{Ours (Err.)} & \textbf{0.135} & \textbf{0.246} & \textbf{0.352} & \textbf{0.191} & \textbf{0.161} \\
			\textbf{Ours (Std.)} & \textbf{0.021} & \textbf{0.039} & \textbf{0.091} & \textbf{0.044} & \textbf{0.040} \\ \midrule[\heavyrulewidth]
			\textbf{Seq.} & \textbf{06} & \textbf{07} & \textbf{09} & \textbf{10} & \textbf{Avg.} \\ \midrule
			VGGT-L \cite{VGGT-Long} (Err.) & 0.964 & 2.581 & 1.682 & 1.858 & 1.402 \\
			VGGT-L \cite{VGGT-Long} (Std.) & 0.567 & 5.088 & 1.506 & 2.962 & 1.643 \\ \midrule
			\textbf{Ours (Err.)} & \textbf{0.218} & \textbf{0.318} & \textbf{0.195} & \textbf{0.331} & \textbf{0.239} \\
			\textbf{Ours (Std.)} & \textbf{0.027} & \textbf{0.070} & \textbf{0.032} & \textbf{0.083} & \textbf{0.050} \\ \bottomrule

		\end{tabular}%
	}
\end{table}

\begin{figure}[t]
	\centering
	\subfloat[Seq. 00]{
		\includegraphics[width=0.95\columnwidth]{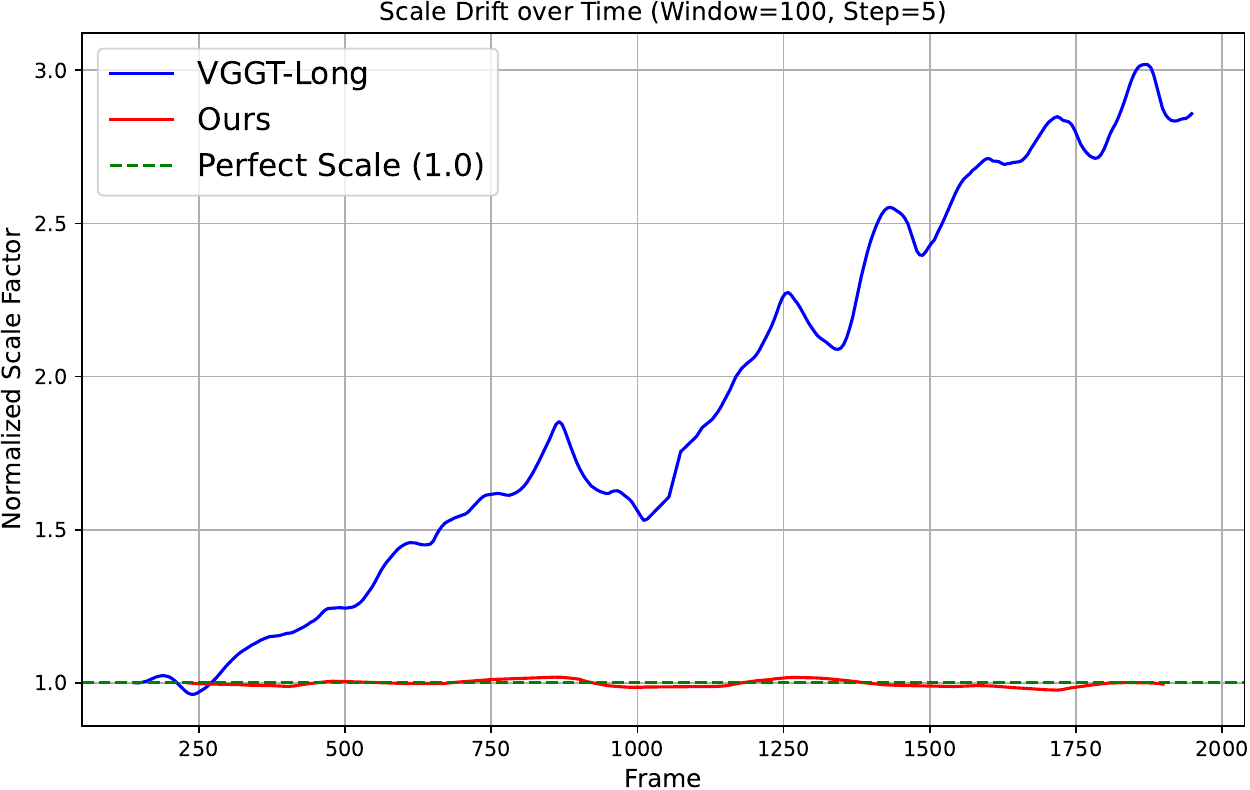}
	}
	\\
	\subfloat[Seq. 05]{
		\includegraphics[width=0.95\columnwidth]{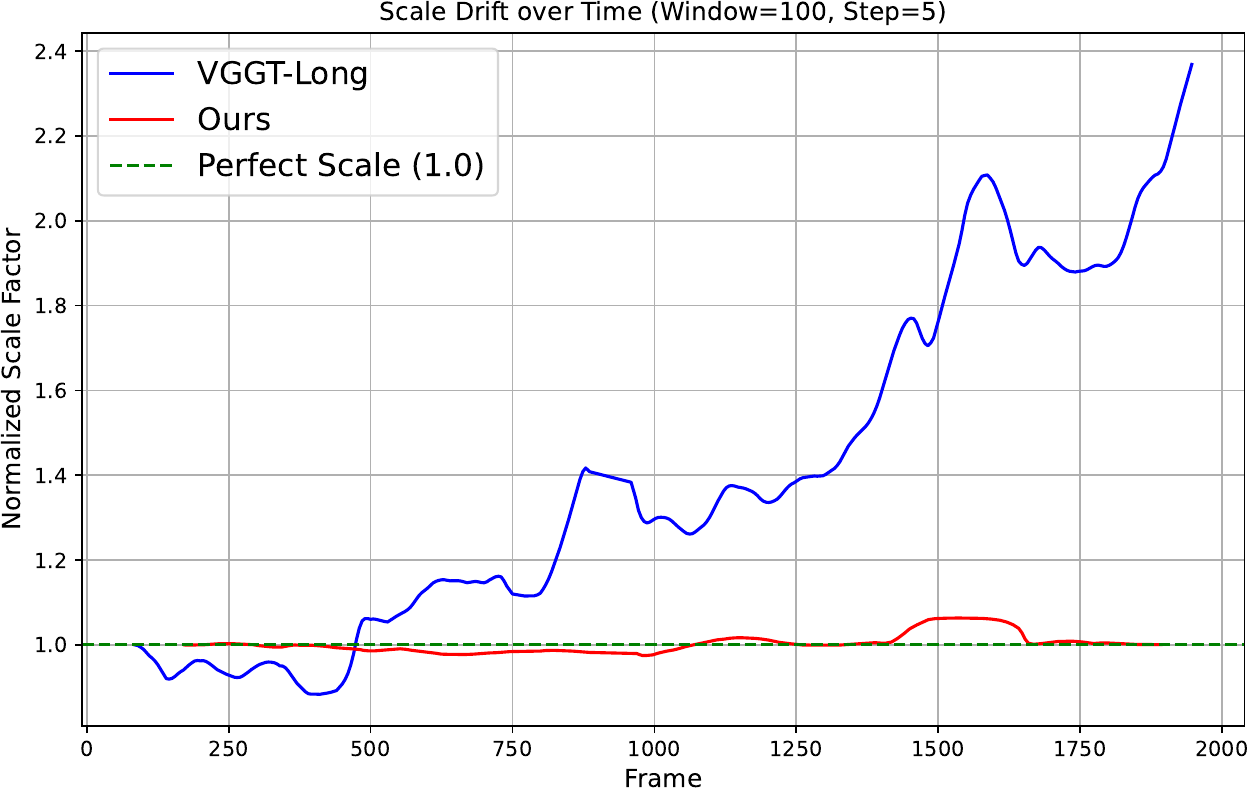}
	}
	\caption{Qualitative comparison of scale drift analysis on KITTI-360 \cite{KITTI-360}. The plots visualize the normalized scale factor evolution over sliding windows for Seq. 00 (top) and Seq. 05 (bottom). The \textcolor{green}{green dashed line} indicates the ideal ground truth scale (1.0). The \textcolor{blue}{blue curve} represents VGGT-Long \cite{VGGT-Long}, which exhibits continuous drift, while the \textcolor{red}{red curve} shows CAL$^\text{2}$M maintaining a stable scale around the ground truth.}
	\label{fig:scale_drift}
\end{figure}

\subsubsection{Scale Consistency Analysis}
We evaluated the scale stability of CAL$^\text{2}$M against VGGT-Long \cite{VGGT-Long} using the KITTI-360 dataset \cite{KITTI-360}. To quantify drift, we employed a sliding window approach: we selected windows of 100 frames with a stride of 5 frames. For each window, we computed the scale factor relative to the ground truth and normalized it by the scale of the first window (ideal normalized scale = 1.0).

Quantitative analysis indicates that CAL$^\text{2}$M achieves significantly lower mean error and standard deviation in scale factors compared to VGGT-Long \cite{VGGT-Long} across all sequences. As illustrated in the qualitative curves (Fig.~\ref{fig:scale_drift}), the scale of VGGT-Long \cite{VGGT-Long} continuously drifts away from the ideal value, whereas our method maintains a stable scale factor fluctuating narrowly around 1.0 throughout the trajectory. This confirms that our assistant-eye constraints effectively eliminate the scale drift characteristic of VGFM extensions.
\subsection{Runtime Analysis}
\label{subsec:runtime}

We analyze the computational efficiency of the CAL$^\text{2}$M framework. The breakdown of the average processing time for our core algorithmic components per sub-map is detailed in Table~\ref{tab:runtime}.
Regarding our proposed contributions, the core processing modules require an average of 1.12s per sub-map. Note that the computationally heavier Test Bank Construction (0.88s) is only triggered intermittently and stops once the bank is full, making its impact on the overall runtime manageable.

Notably, our system architecture is designed for parallel execution. The VGFM inference run asynchronously from the CAL$^\text{2}$M optimization and mapping components. Since the average processing time of our core components (1.12s) is even less than the VGFM inference time (VGGT, 1.24s), our framework operates in parallel without inducing pipeline stalls. Actually, given a sequence of 2,000 frames, the main pipeline of CAL$^\text{2}$M finally achieves an effective frame rate of 26.2 FPS, while the performance bottleneck dictated mainly by the inference speed of the backbone model.


\begin{table}[h]
	\centering
	\caption{Runtime Breakdown of CAL$^\text{2}$M Components per Sub-map (15 keyframes).}
	\label{tab:runtime}
	\setlength{\tabcolsep}{22pt} 
	\begin{tabular}{cc}
		\toprule
		\textbf{Module} & \textbf{Time (s)} \\
		\midrule
		Test Bank Construction$^*$ & 0.88 \\
		Intrinsic Search & 0.02 \\
		Anchor Extraction \& Mapping & 0.95 \\
		Sub-map Alignment & 1e-4 \\ \midrule
		\textbf{Average Pipeline Time Cost} & \textbf{1.12} \\
		\textbf{VGFM Inference (VGGT)} &  \textbf{1.24}\\
		\bottomrule
		\multicolumn{2}{l}{\footnotesize $^*$Executed intermittently (every 5 sub-maps).}
	\end{tabular}
\end{table}

\subsection{Ablation Studies}
\label{subsec:ablation}

\subsubsection{Localization Components}
To validate the contribution of each individual component within the CAL$^\text{2}$M framework, we conducted a comprehensive ablation study on the KITTI-360 dataset \cite{KITTI-360}. We analyze the impact on localization accuracy by selectively disabling specific modules.
Specifically, we compare the full model against five variants: (1) \textit{w/o Optimization}: disabling the backend Joint PGO; (2) \textit{w/o Pose Correction}: disabling both intrinsic search and the pose correction model; (3) \textit{w/o Rotation Correction}: applying only translation correction; (4) \textit{w/o Translation Correction}: applying only rotation correction; and (5) \textit{w/o Scale Rectification}: relying on the raw VGFM scale.

Relevant experimental results are summarized in Table \ref{tab:ablation_localization}, which unequivocally demonstrate the necessity of each proposed module. The removal of the assistant-eye-based scale rectification leads to the most catastrophic performance drop, confirming that resolving monocular scale ambiguity is the fundamental prerequisite for accurate kilometer-level localization. Furthermore, the pose correction module proves critical for accuracy. Specifically, excluding rotation correction causes significant trajectory divergence due to accumulated orientation errors, while removing translation correction notably degrades the metric step accuracy. Finally, while the backend optimization contributes to the ultimate precision, the system maintains competitive performance even without it, validating the robustness of our front-end rectification strategies.

\begin{table}[h]
	\centering
	\caption{Ablation Study of Localization Components on KITTI-360 \cite{KITTI-360}.}
	\label{tab:ablation_loc}
	\renewcommand{\arraystretch}{1.2}
	\setlength{\tabcolsep}{8pt}
	\begin{tabular}{lcccc}
		\toprule
		\multirow{2}{*}{\textbf{Method Variant}} & \multicolumn{2}{c}{\textbf{ATE (m)} $\downarrow$} & \multicolumn{2}{c}{\textbf{ATE Ratio (\%)} $\downarrow$} \\
		\cmidrule(lr){2-3} \cmidrule(lr){4-5}
		& \textbf{Avg} & \textbf{Avg}$^*$ & \textbf{Avg} & \textbf{Avg}$^*$ \\
		\midrule
		\textbf{Ours (Full Model)} & \textbf{19.07} & \textbf{13.36} & \textbf{0.884} & \textbf{0.795} \\
		w/o Optimization & 19.24 & 13.71 & 0.904 & 0.822 \\
		w/o Pose Correction & 28.18 & 24.15 & 1.466 & 1.463 \\
		\quad \textit{-- w/o Rotation Corr.} & 27.93 & 23.79 & 1.440 & 1.432 \\
		\quad \textit{-- w/o Translation Corr.} & 20.07 & 14.36  & 0.942 & 0.857 \\
		w/o Scale Rectification & 45.14 & 35.52 & 2.194 & 2.092 \\
		\bottomrule
		\multicolumn{5}{l}{\footnotesize $^*$Excludes Sequence 07 (high-speed highway).}
	\end{tabular}
	\label{tab:ablation_localization}
\end{table}

\begin{figure}[h]
	\centering
	\subfloat[Seq. 03]{
		\includegraphics[width=0.95\columnwidth]{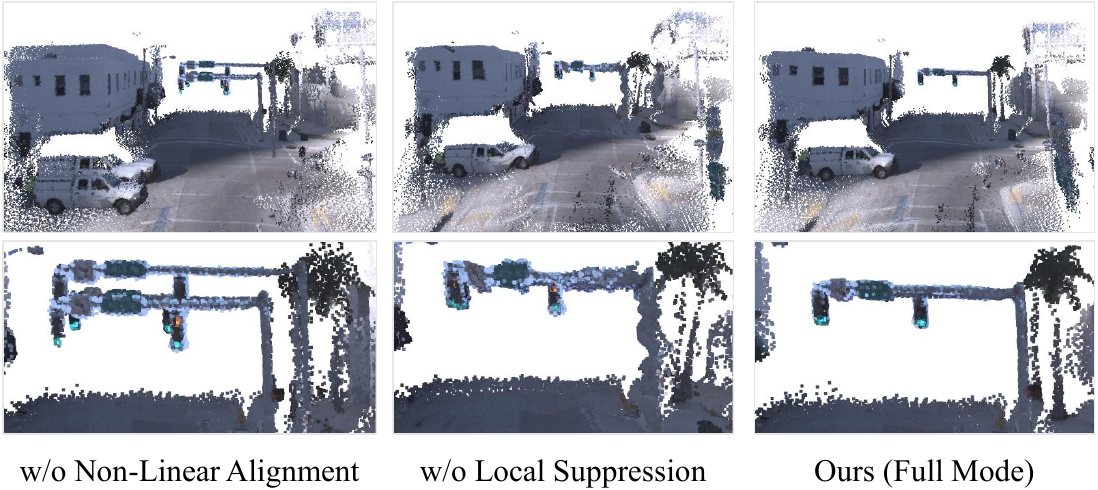}
	}
	\\
	\subfloat[Seq. 05]{
		\includegraphics[width=0.95\columnwidth]{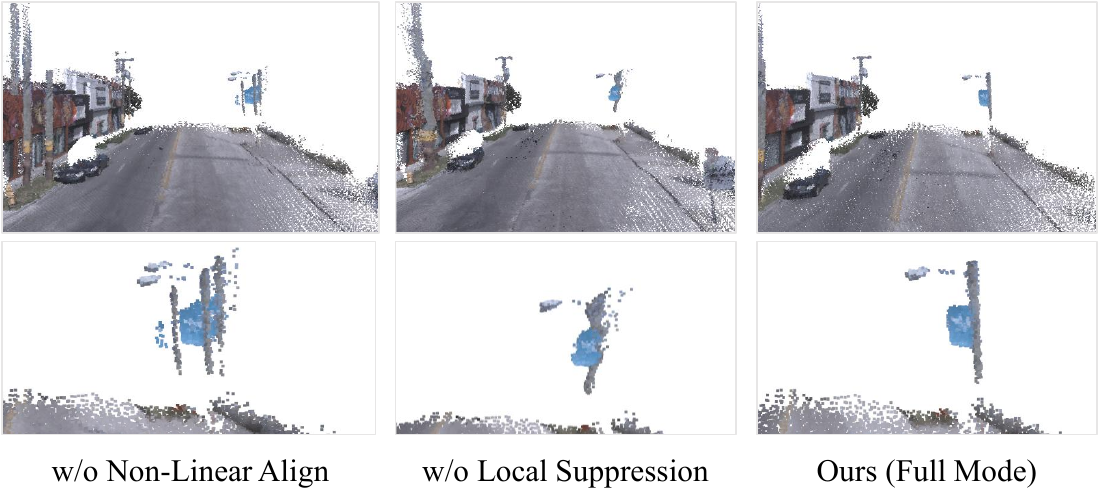}
	}
	\caption{Qualitative ablation study of mapping components on Argoverse \cite{Argoverse} Seq. 03 (top) and Seq. 05 (bottom). Each sub-figure compares three settings: w/o Non-Linear Alignment, w/o Local Suppression, and the Full Model. The upper part of each panel displays the global point cloud, while the lower part provides a zoomed-in view of local details. }
	\label{fig:ablation_map}
\end{figure}

\subsubsection{Mapping Components}
We evaluate the global mapping strategies on the Argoverse dataset \cite{Argoverse} against three variants: (1) \textit{w/o Local Suppression}: disabling the local suppression of anchors; (2) \textit{w/o Adaptive Fusion}: using arithmetic mean for anchor coordinates; and (3) \textit{w/o Non-Linear Alignment}: without using TPS alignment.
Quantitative results in Table~\ref{tab:ablation_map} show that the full model achieves the best Chamfer distance and Accuracy. Although the variant without non-linear alignment yields slightly lower Completeness, this metric is somewhat misleading, since noisy or dispersed point clouds caused by rigid misalignment often artificially improve coverage scores. Therefore, we prioritize Accuracy as the more reliable indicator of geometric quality. The superior Accuracy of our model confirms that our global consistent mapping strategy can effectively correct internal distortions in sub-maps, which is also corroborated by the qualitative results in Fig. \ref{fig:ablation_map}.


\begin{table}[h]
	\centering
	\caption{Ablation Study of Mapping Components on Argoverse \cite{Argoverse}.}
	\label{tab:ablation_map}
	\renewcommand{\arraystretch}{1.2}
	\setlength{\tabcolsep}{8pt}
	\scalebox{0.9}{
	\begin{tabular}{lccc}
		\toprule
		\textbf{Method Variant} & \textbf{Chamfer} $\downarrow$ & \textbf{Accuracy} $\downarrow$ & \textbf{Completeness} $\downarrow$ \\
		\midrule
		\textbf{Ours (Full Model)} & \textbf{1.414} & \textbf{0.919} & 1.909 \\
		w/o Local Suppression & 1.693 & 0.972 & 2.413 \\
		w/o Adaptive Fusion & 1.567 & 0.940 & 2.194 \\
		w/o Non-Linear Align & 1.439 & 0.988 & \textbf{1.890} \\
		\bottomrule
	\end{tabular}}
\vspace{-1mm}
\end{table}

\section{Limitations and Future Work}
\label{sec:limitations}

While CAL$^\text{2}$M represents a significant step toward flexible, calibration-free SLAM, certain limitations still remain. First, our intrinsic search and pose correction module assumes a fixed-focus camera model. Since the method relies on a consistent intrinsic prior to filter estimation noise, this correction strategy becomes ineffective in scenarios involving continuous zoom or variable focal lengths. Extending the test bank mechanism to model time-varying intrinsics represents a promising direction for future research.

Second, our framework prioritizes deployment flexibility by removing strict requirements for calibration and hardware synchronization. Although CAL$^\text{2}$M outperforms calibrated monocular baselines and other calibration-free approaches, a performance gap naturally remains when compared to fully calibrated, synchronized stereo SLAM systems. We view this as an acceptable trade-off for the ability to deploy on arbitrary, loosely coupled dual-camera setups. In future work, we aim to narrow this gap by leveraging the evolution of foundation models and exploring methods to better exploit the constant spacing prior for tighter geometric constraints.

\section{Conclusion}
\label{sec:conclusion}
In this paper, we presented CAL$^\text{2}$M, a robust framework designed to bridge the gap between the local zero-shot capabilities of Visual Geometry Foundation Models and the rigorous demands of kilometer-scale SLAM. By introducing a loosely coupled ``assistant eye'' and leveraging the constant spacing prior, we effectively resolved the inherent scale ambiguity without relying on strict pre-calibration or hardware synchronization. To address the geometric inaccuracies stemming from affine ambiguity, we developed an online intrinsic search and pose correction model rooted in epipolar geometry, which significantly enhances tracking accuracy. Furthermore, our anchor-based global mapping strategy enables elastic, non-linear alignment of sub-maps, ensuring structural consistency across large-scale environments. Extensive evaluations demonstrate that CAL$^\text{2}$M effectively mitigates the drift issues prevalent in existing VGFM extensions, delivering superior accuracy and robustness among calibration-free approaches. We believe our work presents a promising step toward flexible, calibration-free autonomous localization and reconstruction, facilitating the practical deployment of foundation models in real-world scenarios.

%

\begin{IEEEbiography}[{\includegraphics[width=1in,height=1.25in]{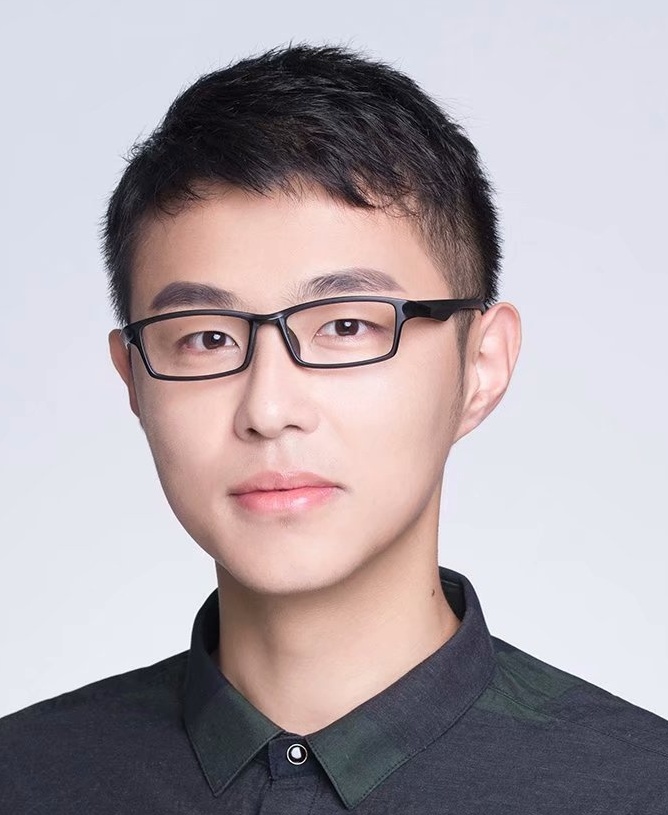}}]{Tianjun Zhang}
	received his B.Eng. and Ph.D. degree from the School of Software Engineering and the School of Computer Science and Technology, Tongji University, Shanghai, China, in 2019 and 2024, respectively. Starting from Dec. 2024, he worked as a postdoctoral at the Department of Automation, Shanghai Jiao Tong University, China. His research interests include collaborative SLAM, computer vision, and sensor calibration.
	\vspace{-0.1cm}
\end{IEEEbiography}

\begin{IEEEbiography}[{\includegraphics[width=1in,height=1.25in]{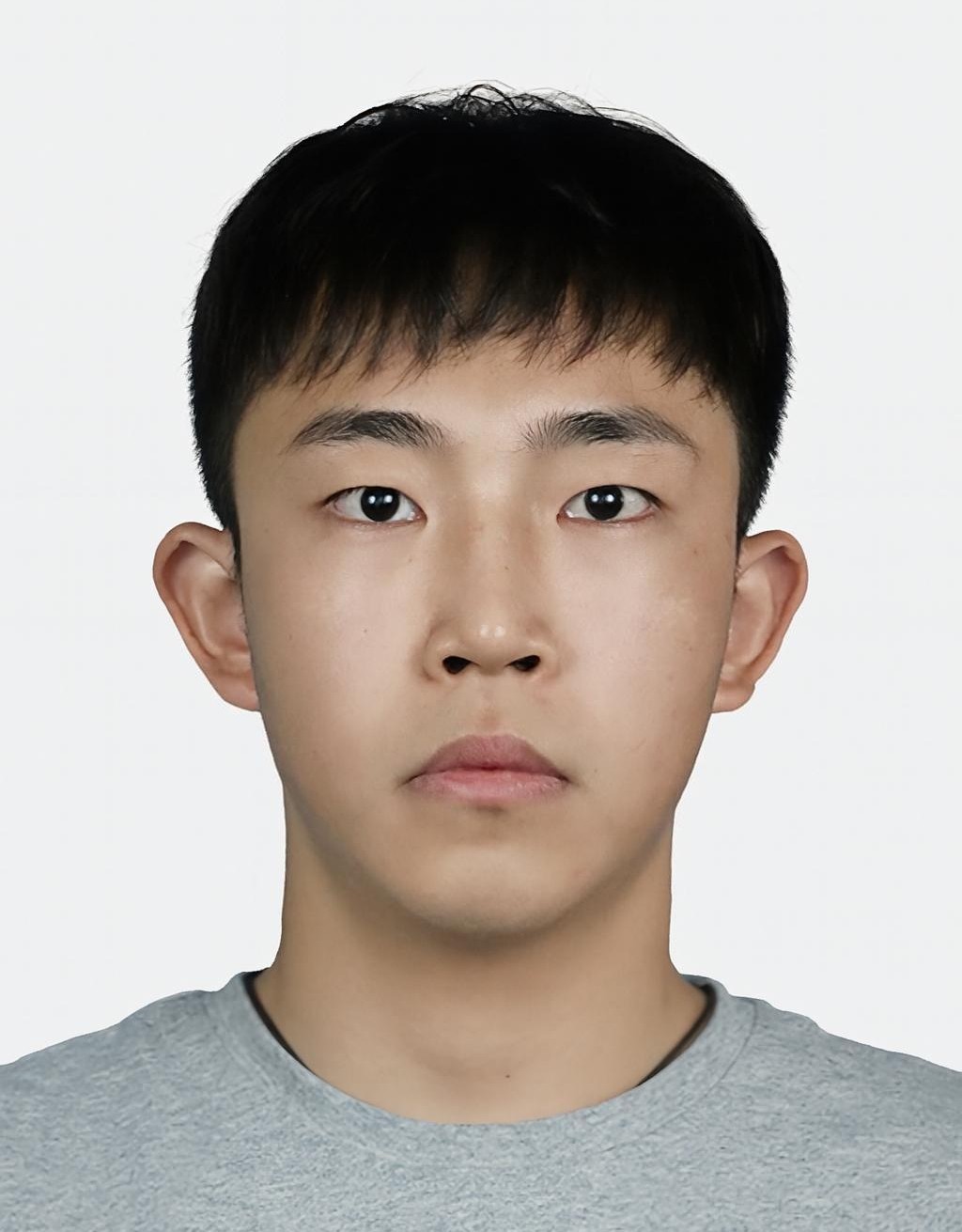}}]{Fengyi Zhang}
	received his B.Eng. degree from the School of Software, Shandong University, Jinan, China, in 2021. He received his M.Eng. degree from the School of Software Engineering, Tongji University, Shanghai, China, in 2024. He is now pursuing his Ph.D. degree at the School of Electrical Engineering and Computer Science, The University of Queensland, Brisbane, Australia. His research focuses on data-driven 3D reconstruction for real-world visual understanding.
	\vspace{-0.1cm}
\end{IEEEbiography}

\begin{IEEEbiography}[{\includegraphics[width=1in,height=1.25in]{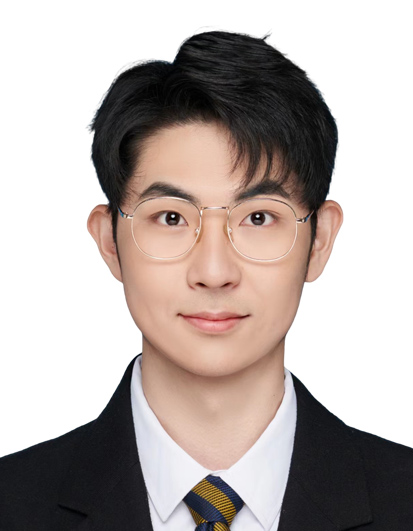}}]{Tianchen Deng}
	(Graduate Student Member, IEEE) received the B.Eng. degree in control science and engineering from Harbin Institute of Technology, Harbin, China, in 2021. He is currently pursuing the joint Ph.D. degree in control science and engineering with Shanghai Jiao Tong University and Nanyang Technological University. His main research interests include visual SLAM, 3D reconstruction, world model, and embodied AI.
	\vspace{-0.1cm}
\end{IEEEbiography}

\begin{IEEEbiography}[{\includegraphics[width=1in,height=1.25in]{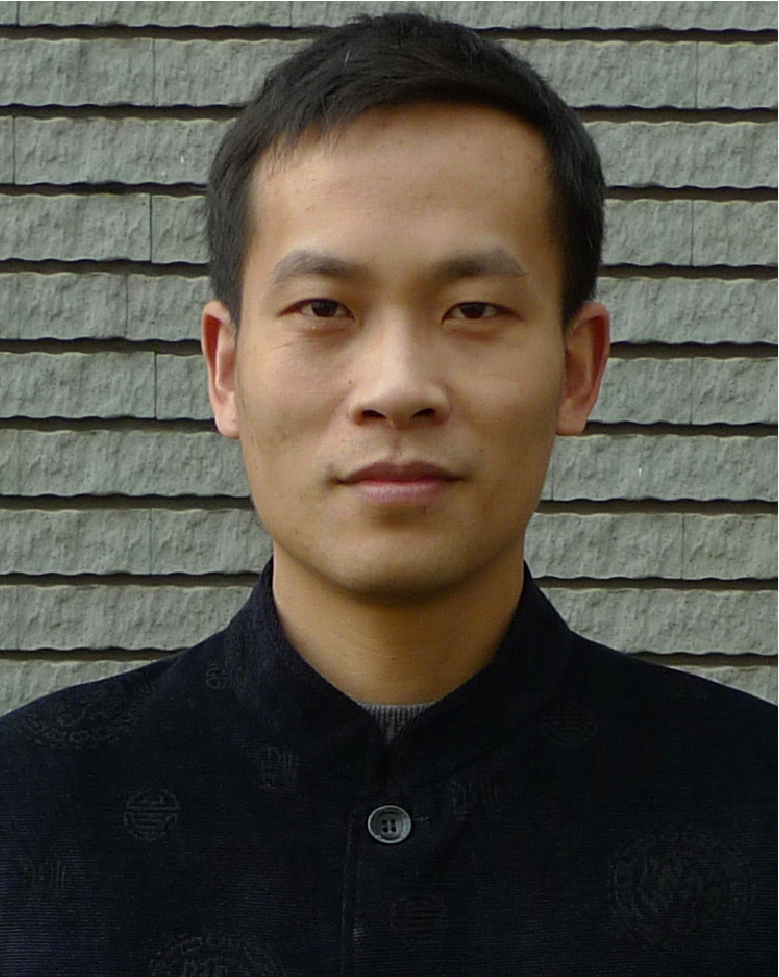}}]{Lin Zhang}
	(Senior Member, IEEE) received the B.Sc. and M.Sc. degrees from the Department of Computer Science and Engineering, Shanghai Jiao Tong University, Shanghai, China, in 2003 and 2006, respectively. He received the Ph.D. degree from the Department of Computing, The Hong Kong Polytechnic University, Hong Kong, in 2011. From March 2011 to August 2011, he was a Research Associate with the Department of Computing, The Hong Kong Polytechnic University. In Aug. 2011, he joined the School of Software Engineering, Tongji University, Shanghai, China. He is currently a Full Professor at the School of Computer Science and Technology, Tongji University. His current research interests include environment perception of intelligent vehicle, pattern recognition, computer vision, and perceptual image/video quality assessment. He serves as an Associate Editor for IEEE Robotics and Automation Letters, and Journal of Visual Communication and Image Representation. He was awarded as a Young Scholar of Changjiang Scholars Program, Ministry of Education, China.
	\vspace{-0.1cm}
\end{IEEEbiography}

\begin{IEEEbiography}[{\includegraphics[width=1in,height=1.25in]{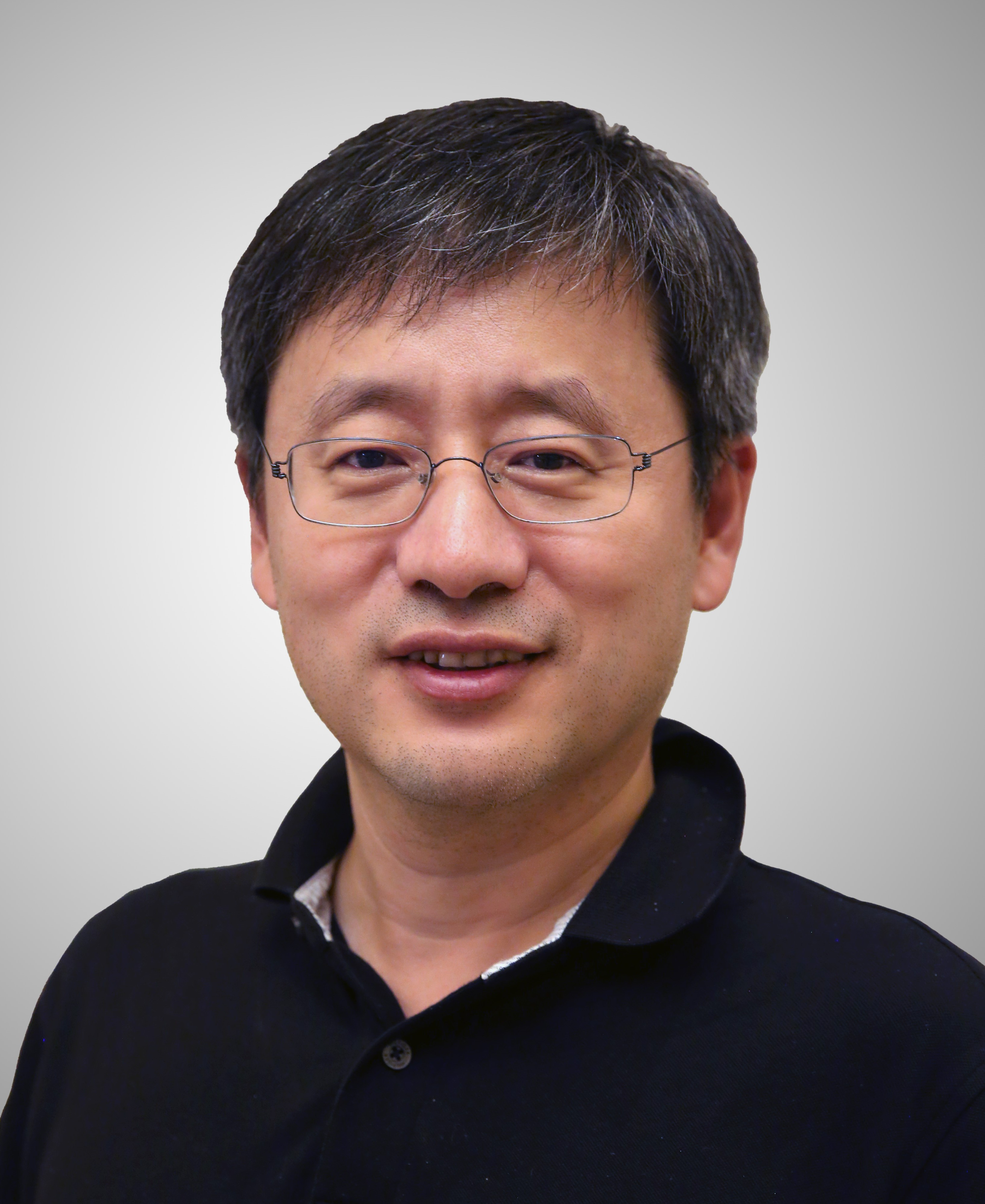}}]{Hesheng Wang} (Senior Member, IEEE) received the B.Eng. degree in electrical engineering from the Harbin Institute of Technology, Harbin, China, in 2002, and the M.Phil. and Ph.D. degrees in automation and computer-aided engineering from The Chinese University of Hong Kong, Hong Kong, in 2004 and 2007, respectively. He is currently a Distinguished Professor with the School of Automation and Intelligent Sensing, Shanghai Jiao Tong University, Shanghai, China. His current research interests include visual servoing, intelligent robotics, computer vision, and autonomous driving. 
Dr. Wang is an Associate Editor of Robotic Intelligence and Automation and the International Journal of Humanoid Robotics, a Senior Editor of the IEEE/ASME Transactions on Mechatronics, an Editor-in-chief of Robot Learning. He served as an Associate Editor of the IEEE Transactions on Robotics from 2015 to 2019, an IEEE Transactions on Automation Science and Engineering from 2021 to 2023, He was the General Chair of IEEE/RSJ IROS 2025, IEEE ROBIO 2022 and IEEE RCAR 2016.
	\vspace{-0.1cm}
\end{IEEEbiography}


\begin{thebibliography}{1}
	
	\bibitem{TPAMI25_UAD}
	M. Guo, Z. Zhang, Y. He,\emph{et al.}, ``End-to-end autonomous driving without costly modularization and 3D manual annotation,'' \emph{IEEE Trans. Pattern Anal. Mach. Intell. (TPAMI)}, Early Access, 2025.
	
	\bibitem{TPAMI25_RoboBEV}
	S. Xie, L. Kong, W. Zhang, \emph{et al.}, ``Benchmarking and improving bird's eye view perception robustness in autonomous driving,'' \emph{IEEE Trans. Pattern Anal. Mach. Intell. (TPAMI)}, vol. 47, no. 5, pp. 3878-3894, 2025.
	
	
	
	\bibitem{ORB-SLAM}
	R. Mur-Artal, J. M. M. Montiel, and J. D. Tard{\'o}s, ``ORB-SLAM: A versatile and accurate monocular SLAM system,'' \emph{IEEE Trans. Robot. (TRO)}, vol. 31, no. 5, pp. 1147-1163, 2015.
	
	
	\bibitem{ORB-SLAM2}
	R. Mur-Artal and J. D. Tard{\'o}s, ``ORB-SLAM2: An open-source SLAM system for monocular, stereo, and RGB-D cameras,'' \emph{IEEE Trans. Robot. (TRO)}, vol. 33, no. 5, pp. 1255-1262, 2017.
	
	\bibitem{ORB-SLAM3}
	C. Campos, R. Elvira, J. J. G. Rodr{\'i}guez, \emph{et al.}, ``ORB-SLAM3: An accurate open-source library for visual, visual–inertial, and multimap SLAM,'' \emph{IEEE Trans. Robot. (TRO)}, vol. 37, no. 6, pp. 1874-1890, 2021.
	
	\bibitem{SVO}
	C. Forster, M. Pizzoli, and D. Scaramuzza, ``SVO: Fast semi-direct monocular visual odometry,'' in \emph{Proc. IEEE Int'l Conf. Robot. Autom. (ICRA)}, 2014, pp. 15-22.
	
	\bibitem{LSD-SLAM}
	J. Engel, T. Sch{\"o}ps, and D. Cremers, ``LSD-SLAM: Large-scale direct monocular SLAM,'' in \emph{Proc. Eur. Conf. Comput. Vis. (ECCV)}, 2014, pp. 834-849.
	
	\bibitem{DSO}
	J. Engel, V. Koltun, and D. Cremers, ``Direct sparse odometry,'' \emph{IEEE Trans. Pattern Anal. Mach. Intell. (TPAMI)}, vol. 40, no. 3, pp. 611-625, 2017.
	
	
	
	\bibitem{LDSO}
	X. Gao, R. Wang, N. Demmel, \emph{et al.}, ``LDSO: Direct sparse odometry with loop closure,'' in \emph{ Proc. IEEE/RSJ Int'l Conf. Intell. Robots and Syst. (IROS)}, 2018, pp. 2198-2204.
	
	
	\bibitem{Autonomous}
	P. Zhang, Y. Cheng, X. Sun, \emph{et al.}, ``A step toward world models: A survey on robotic manipulation,''  \emph{arXiv preprint arXiv:2511.02097}, 2025.
	
	\bibitem{TPAMI25_MBASLAM}
	P. Wang, L. Zhao, Y. Zhang, \emph{et al.} ``MBA-SLAM: Motion blur aware dense visual SLAM with radiance fields representation,'' \emph{IEEE Trans. Pattern Anal. Mach. Intell. (TPAMI)}, Early Access, 2025.
	
	\bibitem{TPAMI25_LN3Diff}
	Y. Lan, F. Hong, S. Zhou, \emph{et al.}, ``LN3Diff++: Scalable latent neural fields diffusion for speedy 3D generation,'' \emph{IEEE Trans. Pattern Anal. Mach. Intell. (TPAMI)}, Early Access, 2025.
	
	
		\bibitem{DUSt3R}
	S. Wang, V. Leroy, Y. Cabon, \emph{et al.}, ``DUSt3R: Geometric 3D vision made easy,'' in \emph{Proc. IEEE/CVF Conf. Comput. Vis. Pattern Recognit. (CVPR)}, 2024, pp. 20697-20709.
	
	\bibitem{Mast3R}
	V. Leroy, Y. Cabon, and J. Revaud, ``Grounding image matching in 3D with MASt3R,'' in \emph{Proc. Eur. Conf. Comput. Vis. (ECCV)}, 2024, pp. 71-91.
	
	\bibitem{VGGT}
	J. Wang, M. Chen, N. Karaev, \emph{et al.}, ``VGGT: Visual geometry grounded transformer,'' in \emph{Proc. IEEE/CVF Conf. Comput. Vis. Pattern Recognit. (CVPR)}, 2025, pp. 5294-5306.
	
	
	\bibitem{Pi3}
	Y. Wang, J. Zhou, H. Zhu, \emph{et al.}, ``$\pi^ 3$: Permutation-equivariant visual geometry learning,'' \emph{arXiv preprint arXiv:2507.13347}, 2025.
	
	
	
	\bibitem{VGGT-Long}
	K. Deng, Z. Ti, J. Xu, \emph{et al.}, ``VGGT-Long: Chunk it, loop it, align it--pushing VGGT's limits on kilometer-scale long RGB sequences,'' \emph{arXiv preprint arXiv:2507.16443}, 2025.
	
	\bibitem{VGGT-SLAM}
	D. Maggio, H. Lim, and L. Carlone, ``VGGT-SLAM: Dense RGB SLAM optimized on the SL(4) manifold,'' \emph{arXiv preprint arXiv:2505.12549}, 2025.
	
	\bibitem{Colmap1}
	J. L. Sch{\"o}nberger and J.M. Frahm, ``Structure-from-motion revisited,'' in \emph{Proc. IEEE Conf. Comput. Vis. Pattern Recognit. (CVPR)}, 2016, pp. 4104-4113.
	
	\bibitem{Colmap2}
	J. L. Sch{\"o}nberger, E. Zheng, J. M. Frahm, \emph{et al.}, ``Pixelwise view selection for unstructured multi-view stereo,'' in \emph{Proc. Eur. Conf. Comput. Vis. (ECCV)}, 2016, pp. 501-518.
	
	\bibitem{SIFT}
	D. G. Lowe, ``Distinctive image features from scale-invariant keypoints,'' \emph{Int'l J. Comput. Vis. (IJCV)}, vol. 60, no. 2, pp. 91-110, 2004.
	
	
	\bibitem{SGM}
	H. Hirschmuller, ``Stereo processing by semiglobal matching and mutual information,'' \emph{IEEE Trans. Pattern Anal. Mach. Intell. (TPAMI)}, vol. 30, no. 2, pp. 328-341, 2008.
	
	\bibitem{Furukawa2010Stereo}
	Y. Furukawa and J. Ponce, ``Accurate, dense, and robust multiview stereopsis,'' \emph{IEEE Trans. Pattern Anal. Mach. Intell. (TPAMI)}, vol. 32, no. 8, pp. 1362-1376, 2009.
	
	
	\bibitem{MVSNet}
	Y. Yao, Z. Luo, S. Li, \emph{et al.}, ``MVSNet: Depth inference for unstructured multi-view stereo,'' in \emph{Proc. Eur. Conf. Comput. Vis. (ECCV)}, 2018, pp. 767-783.
	
	\bibitem{RecurrentMVSNet}
	Y. Yao, Z. Luo, S. Li, \emph{et al.}, ``Recurrent MVSNet for high-resolution multi-view stereo depth inference,'' in \emph{Proc. IEEE/CVF Conf. Comput. Vis. Pattern Recognit. (CVPR)}, 2019, pp. 5525-5534.
	
	
	\bibitem{Gu2020Multiview}
	X. Gu, Z. Fan, S. Zhu, \emph{et al.}, ``Cascade cost volume for high-resolution multi-view stereo and stereo matching,'' in \emph{Proc. IEEE/CVF Conf. Comput. Vis. Pattern Recognit. (CVPR)}, 2020, pp. 2495-2504.
	
	\bibitem{PatchMatchNet}
	F. Wang, S. Galliani, C. Vogel, \emph{et al.}, ``PatchmatchNet: Learned multi-view patchmatch stereo,'' in \emph{Proc. IEEE/CVF Conf. Comput. Vis. Pattern Recognit. (CVPR)}, 2021, pp. 14194-14203.
	
	\bibitem{NeRF}
	B. Mildenhall, P. P. Srinivasan, M. Tancik, \emph{et al.}, ``NeRF: Representing scenes as neural radiance fields for view synthesis,'' in \emph{Proc. Eur. Conf. Comput. Vis. (ECCV)}, 2020, pp. 405-421.
	
	\bibitem{InstantNGP}
	T. M{\"u}ller, A. Evans, C. Schied, \emph{et al.}, ``Instant neural graphics primitives with a multiresolution hash encoding,'' \emph{ACM Trans. Graph. (TOG)}, vol. 41, no. 4, pp. 1-15, 2022.
	
	\bibitem{MipNeRF}
	J. T. Barron, B. Mildenhall, M. Tancik, \emph{et al.}, ``Mip-NeRF: A multiscale representation for anti-aliasing neural radiance fields,'' in \emph{Proc. IEEE/CVF int'l Conf. Comput. Vis. (ICCV)}, 2021, pp. 5855-5864.
	
	\bibitem{PixelNeRF}
	A. Yu, V. Ye, M. Tancik, \emph{et al.}, ``PixelNeRF: Neural radiance fields from one or few images,'' in \emph{Proc. IEEE/CVF Conf. Comput. Vis. Pattern Recognit. (CVPR)}, 2021, pp. 4578-4587.
	
	\bibitem{IBRNet}
	Q. Wang, Z. Wang, K. Genova, \emph{et al.}, ``IBRNet: Learning multi-view image-based rendering,'' in \emph{Proc. IEEE/CVF Conf. Comput. Vis. Pattern Recognit. (CVPR)}, 2021, pp. 4690-4699.
	
	\bibitem{MVSNeRF}
	A. Chen, Z. Xu, F. Zhao, \emph{et al.}, ``MVSNeRF: Fast generalizable radiance field reconstruction from multi-view stereo,'' in \emph{Proc. IEEE/CVF Int'l Conf. Comput. Vis. (ICCV)}, 2021, pp. 14124-14133.
	
	\bibitem{3DGS}
	B. Kerbl, G. Kopanas, T. Leimk{\"u}hler, \emph{et al.}, ``3D Gaussian splatting for real-time radiance field rendering,'' \emph{ACM Trans. Graph. (TOG)}, vol. 42, no. 4, pp. 139:1-139:14, 2023.
	
%
	
	
	\bibitem{MonoSLAM}
	A. J. Davison, I. D. Reid, N. D. Molton, \emph{et al.}, ``MonoSLAM: Real-time single camera SLAM,'' \emph{IEEE Trans. Pattern Anal. Mach. Intell. (TPAMI)}, vol. 29, no. 6, pp. 1052-1067, 2007.
	
	\bibitem{PTAM}
	G. Klein and D. Murray, ``Parallel tracking and mapping for small AR workspaces,'' in \emph{Proc. IEEE/ACM Int'l Symp. Mixed Augmented Reality (ISMAR)}, 2007, pp. 225-234.
	
	
	\bibitem{SPTAM}
	T. Pire, T. Fischer, G. Castro, \emph{et al.}, ``S-PTAM: Stereo parallel tracking and mapping,'' \emph{Robot. Auton. Syst. (RAS)}, vol. 93, pp. 27-42, 2017.
	
	
	
	\bibitem{StereoLSD}
	J. Engel, J. St{\"u}ckler, and D. Cremers, ``Large-scale direct SLAM with stereo cameras,'' in \emph{Proc. IEEE/RSJ Int'l Conf. Intell. Robots Syst. (IROS)}, 2015, pp. 1935-1942.
	
	\bibitem{DROID-SLAM}
	Z. Teed and J. Deng, ``DROID-SLAM: Deep visual SLAM for monocular, stereo, and RGB-D cameras,'' in \emph{Proc. Adv. Neural Inf. Process. Syst. (NeurIPS)}, 2021, pp. 16558-16569.
	
	\bibitem{DPVO}
	Z. Teed, L. Lipson, and J. Deng, ``Deep patch visual odometry,'' in \emph{Proc. Adv. Neural Inf. Process. Syst. (NeurIPS)}, 2023, pp. 39033-39051.
	
	\bibitem{DPV-SLAM}
	L. Lipson, Z. Teed, and J. Deng, ``Deep patch visual SLAM,'' in \emph{Proc. Eur. Conf. Comput. Vis. (ECCV)}, 2024, pp. 424-440.
	
	\bibitem{TartanVO}
	W. Wang, Y. Hu, and S. Scherer, ``TartanVO: A generalizable learning-based VO,'' in \emph{Proc. Conf. Robot Learning (CoRL)}, 2021, pp. 1761-1772.
	
	\bibitem{IMAP}
	E. Sucar, S. Liu, J. Ortiz, \emph{et al.} ``iMap: Implicit mapping and positioning in real-time,'' in \emph{Proc. IEEE/CVF Int'l Conf. Computer Vis. (ICCV)}. 2021, pp. 6229-6238.
	
	
	\bibitem{NICE-SLAM}
	Z. Zhu, S. Peng, V. Larsson, \emph{et al.}, ``NICE-SLAM: Neural implicit scalable encoding for slam,'' in \emph{Proc. IEEE/CVF Conf. Computer Vis. Pattern Recognit. (CVPR)}, 2022, pp. 12786-12796.
	
	
	\bibitem{LoopySLAM}
	L. Liso, E. Sandstr{\"o}m, V. Yugay, \emph{et al.}, ``Loopy-SLAM: Dense neural SLAM with loop closures,'' in \emph{Proc. IEEE/CVF Conf. Comput. Vis. Pattern Recognit. (CVPR)}, 2024, pp. 20363-20373.
	
	\bibitem{deng2025mne}
	T. Deng, G. Shen, C. Xun, \emph{et al.}, ``MNE-SLAM: Multi-agent neural SLAM for mobile robots,'' in \emph{Proc. IEEE/CVF Conf. Comput. Vis. Pattern Recognit. (CVPR)}, 2025, pp. 1485-1494.
	
	\bibitem{SplaTAM}
	N. Keetha, J. Karhade, K. M. Jatavallabhula, \emph{et al.}, ``SplaTAM: Splat, track \& map 3D Gaussians for dense RGB-D SLAM,'' in \emph{Proc. IEEE/CVF Conf. Comput. Vis. Pattern Recognit. (CVPR)}, 2024, pp. 21357-21366.
	
	\bibitem{GS-SLAM}
	C. Yan, D. Qu, D. Xu, \emph{et al.}, ``GS-SLAM: Dense visual SLAM with 3D Gaussian splatting,'' in \emph{Proc. IEEE/CVF Conf. Comput. Vis. Pattern Recognit. (CVPR)}, 2024, pp. 19595-19604.
	
	\bibitem{MonoGS}
	H. Matsuki, R. Murai, P. H. J. Kelly, \emph{et al.}, ``Gaussian splatting SLAM,'' in \emph{Proc. IEEE/CVF Conf. Comput. Vis. Pattern Recognit. (CVPR)}, 2024, pp. 18039-18048.
	
	
	
	\bibitem{Fast3R}
	J. Yang, A. Sax, K. J. Liang, \emph{et al.}, ``Fast3R: Towards 3D reconstruction of 1000+ images in one forward pass,'' in \emph{Proc. IEEE/CVF Conf. Comput. Vis. Pattern Recognit. (CVPR)}, 2025, pp. 21924-21935.
	
	\bibitem{MUSt3R}
	Y. Cabon, L. Stoffl, L. Antsfeld, \emph{et al.}, ``MUSt3R: Multi-view network for stereo 3D reconstruction,'' in \emph{Proc. IEEE/CVF Conf. Comput. Vis. Pattern Recognit. (CVPR)}, 2025, pp. 1050-1060.
	
	
	\bibitem{deng2025relocvggtvisualrelocalizationgeometry}
	T. Deng, W. Wu, K. Wu, \emph{et al.}, ``Reloc-VGGT: Visual re-localization with geometry grounded Transformer,'' \emph{arXiv preprint arXiv:2512.21883}, 2025.
	
	
	\bibitem{unipr-3d}
	T. Deng, X. Chen, Z. Liu, \emph{et al.}, ``UniPR-3D: Towards universal visual place recognition with visual geometry grounded Transformer,'' \emph{arXiv preprint arXiv:2512.21078}, 2025.
	
	\bibitem{deng2025best3dscenerepresentation}
	T. Deng, Y. Pan, S. Yuan, \emph{et al.}, ``What is the best 3D scene representation for robotics? From geometric to foundation models,'' \emph{arXiv preprint arXiv:2512.03422}, 2025.
	

	
	\bibitem{MapAnything}
	N. Keetha, N. M{\"u}ller, J. Sch{\"o}nberger, \emph{et al.}, ``MapAnything: Universal feed-forward metric 3D reconstruction,'' \emph{arXiv preprint arXiv:2509.13414}, 2025.
	
	\bibitem{Span3R}
	H. Wang and L. Agapito, ``3D reconstruction with spatial memory,'' in \emph{Proc. Int'l Conf. 3D Vision (3DV)}, 2025, pp. 78-89.
	
	\bibitem{SLAM3R}
	Y. Liu, S. Dong, S. Wang, \emph{et al.}, ``SLAM3R: Real-time dense scene reconstruction from monocular RGB videos,'' in \emph{Proc. IEEE/CVF Conf. Comput. Vis. Pattern Recognit. (CVPR)}, 2025, pp. 16651-16662.
	
	\bibitem{Mast3rSLAM}
	R. Murai, E. Dexheimer, and A. J. Davison, ``MASt3R-SLAM: Real-time dense SLAM with 3D reconstruction priors,'' in \emph{Proc. IEEE/CVF Conf. Comput. Vis. Pattern Recognit. (CVPR)}, 2025, pp. 16695-16705.
	
	\bibitem{CUT3R}
	Q. Wang, Y. Zhang, A. Holynski, \emph{et al.} ``Continuous 3d perception model with persistent state," in \emph{Proc. Computer Vis. Pattern Recognit. (CVPR)}, 2025, pp. 10510-10522.
	
	\bibitem{ShiTomasi}
	J. Shi and C. Tomasi, ``Good features to track,'' in \emph{Proc. IEEE Conf. Computer Vis. Pattern Recognit. (CVPR)}, 1994, pp. 593-600.
	
	\bibitem{LK}
	B. D. Lucas and T. Kanade, ``An iterative image registration technique with an application to stereo vision,'' in \emph{Proc. Int'l Joint Conf. Artif. Intell. (IJCAI)}, 1981, pp. 674-679.
	
	
	\bibitem{GTSAM}
	F. Dellaert, and GTSAM Contributors, ``Borglab/GTSAM,'' \emph{https://github.com/borglab/gtsam}, 2022.
	
	
	\bibitem{SALAD}
	S. Izquierdo, J. Civera. ``Optimal transport aggregation for visual place recognition,'' in \emph{Proc. IEEE/CVF Conf. Computer Vis. Pattern Recognit. (CVPR)}, 2024, pp. 17658-17668.
	
	\bibitem{TPS}
	F. L. Bookstein, ``Principal warps: thin-plate splines and the decomposition of deformations,'' \emph{IEEE Trans. Pattern Anal. Mach. Intell. (TPAMI)}, vol. 11, no. 6, pp. 567-585, 1989.
	
	\bibitem{KITTI-Odom}
	A. Geiger, P. Lenz, R. Urtasun, ``Are we ready for autonomous driving? The KITTI vision benchmark suite,'' in \emph{Proc. IEEE Conf. Computer Vis. Pattern Recognit. (CVPR)}, 2012, pp. 3354-3361.
	
	\bibitem{KITTI-360}
	Y. Liao, J. Xie, A. Geiger, ``KITTI-360: A novel dataset and benchmarks for urban scene understanding in 2d and 3d,'' \emph{IEEE Trans. Pattern Anal. Mach. Intell. (TPAMI)}, 2022, vol. 45, no. 3, pp. 3292-3310.
	
	\bibitem{Argoverse}
	M. F. Chang, J. Lambert, P. Sangkloy, \emph{et al.}, ``Argoverse: 3d tracking and forecasting with rich maps,'' in \emph{Proc. IEEE/CVF Conf. Computer Vis. Pattern Recognit. (CVPR)}, 2019, pp. 8748-8757.
	
\end{thebibliography}
\end{document}